\documentclass[10pt,journal,cspaper,compsoc]{IEEEtran}
\usepackage{epsfig}
\usepackage{graphicx}
\usepackage{mathtools}
\usepackage{amsmath}
\usepackage{amssymb}
\usepackage{epstopdf}
\usepackage{subfigure}
\usepackage{indentfirst} 
\usepackage{times}
\usepackage{epsfig}
\usepackage{graphicx}
\usepackage{amsmath}
\usepackage{amssymb}
\usepackage{hyperref}
\usepackage{url}
\usepackage{epstopdf}
\usepackage{indentfirst} 

\newcommand{\etal}{\textit{et al.}}
\newtheorem{proposition}{Proposition}
\newcommand{\norm}[1]{\left\lVert#1\right\rVert}
\newcommand{\argmin}{\operatornamewithlimits{arg\ min}}
\ifCLASSOPTIONcompsoc
\else
\fi
\ifCLASSINFOpdf
\else
\fi

\begin{document}
	%
	\title{Fast Integral Image Estimation at 1\% measurement rate}
	%
	%
	%
	%
	
	\author{Kuldeep~Kulkarni,
		Pavan~Turaga
		\IEEEcompsocitemizethanks{\IEEEcompsocthanksitem K. Kulkarni and P. Turaga are with the School of Arts, Media and Engineering and School of Electrical, Computer and Energy Engineering, Arizona State University. Email: kkulkar1@asu.edu, pturaga@asu.edu.\protect\\
		}
		\thanks{}}
	
	%
	%


	\IEEEcompsoctitleabstractindextext{%
	\begin{abstract}
		We propose a framework called {\bf ReFInE} to directly obtain integral image estimates from a very small number of spatially multiplexed measurements of the scene without iterative reconstruction of any auxiliary image, and demonstrate their practical utility in visual object tracking. Specifically, we design measurement matrices which are tailored to facilitate extremely fast estimation of the integral image, by using a single-shot linear operation on the measured vector. Leveraging a prior model for the images, we formulate a nuclear norm minimization problem with second order conic constraints to jointly obtain the measurement matrix and the linear operator. Through qualitative and quantitative experiments, we show that high quality integral image estimates can be obtained using our framework at very low measurement rates. Further, on a standard dataset of 50 videos, we present object tracking results which are comparable to the state-of-the-art methods, even at an extremely low measurement rate of 1\%.
	\end{abstract}
		
		}

		\maketitle

		\IEEEdisplaynotcompsoctitleabstractindextext
		
		\IEEEpeerreviewmaketitle
	
		\section{Introduction}
		In this paper, we study the problem of obtaining integral image estimates from emerging flexible programmable imaging devices. These novel imaging devices, often considered under the broad umbrella of spatial-multiplexing cameras (SMCs)\cite{nayar2006programmable,SPC,sankaranarayanan2012cs} provide a number of benefits in reducing the amount of sensing for portable and resource constrained acquisition. The imaging architectures in these cameras employ spatial light modulators like digital micromirror arrays to optically compute projections of the scene. Mathematically, the projections are given by $y = \phi x$, where $x \in \mathbb{R}^n$ is the image, $y \in \mathbb{R}^m$, known as the measurement vector, denotes the set of sensed projections and $\phi \in \mathbb{R}^{m \times n}$ is called measurement matrix defined by the set of multiplexing patterns. The nature of  the acquisition framework enables us to deploy SMCs in resource constrained settings, wherein one can employ $m << n$ number of photon detectors to sense otherwise high-resolution imagery \cite{SPC,CSLDS} and obtain a very small number of measurements. Later, a reconstruction algorithm is used to recover the image $x$. However, reconstructing $x$ from $y$ when $m < n$ is an ill-posed problem. Researchers in the past have attempted to provide solutions by carefully designing the measurement matrix $\phi$ in the hope of easier recovery of $x$ from $y$. Recent compressive sensing (CS) theory provides one possible solution to tackle the above mentioned ill-posed problem. According to CS theory, a signal can be recovered perfectly from a small number of $m$ $=$ $\mathcal{O}$($s$ log($\frac{n}{s}$)) such pseudo-random (PR) multiplexed measurements, where $s$ is the sparsity of the signal. However, a significant research shows that high-quality reconstruction is computationally intensive \cite{donoho2006compressed,candes2006near,tropp2007signal,needell2009cosamp}. Hence, despite the promise of CS-based SMCs \cite{SPC,sankaranarayanan2012cs}, the computational bottleneck of non-linear, iterative reconstruction has withheld their wide-spread adoption in applications which require fast inference of objects, actions, and scenes. This has led to researchers exploring the option of tackling inference problems directly from these pseudo-random multiplexed measurements \cite{CSLDS,EPFL,Rectex,Davenport1,CLearning,kulkarni2015reconstruction,lohit2015reconstruction} (more on these later in the section in related work). However, the `universal' nature of such measurements has made it challenging to devise new or adopt existing computer vision algorithms to solve the inference problem at hand.
		\noindent
		\par 
		The need to acquire as less data as possible combined with the limitations of pseudo-random multiplexers props us to outline the following goal. The goal of this paper is to propose a novel sensing framework for SMCs such that acquired measurements satisfy the following properties. 1) The measurements are not random in nature but are tailored for a particular application. 2) The number of measurements is 2 orders less than the number of pixels, so that SMCs based on our framework can be employed in resource constrained applications. 3) A simple linear operation on the measurement vector $y$ yields a `proxy' representation (e.g integral images, gradient images) from which the required features are extracted for the application in hand, thus avoiding the computationally expensive iterative and non-linear reconstruction.
		\noindent
		\par
		In this paper, we focus on one such `proxy' representation, integral images. Integral images are extremely attractive representation since Haar-like features and box-filtered image outputs can be computed from integral images with a small and fixed number of  floating point operations in constant time \cite{viola2004robust}. These advantages have led to their widespread use in real time applications like face detection \cite{viola2004robust}, pedestrian detection \cite{dollar2010fastest}, object tracking \cite{grabner2006real,zhang2012real,babenko2011robust,kalal2010pn} and  object segmentation \cite{ramakanth2013seamseg}. 
		
		\noindent
		\par
		Instead of setting a fixed number of measurements, we formulate an optimization problem to minimize the number of measurements while incorporating the other two (1 and 3) properties of measurements (as mentioned above) in the constraints. Minimizing the number of measurements is akin to minimizing the rank of the measurement matrix. In more concrete terms, the problem is posed to jointly minimize the rank of the measurement matrix, $\phi$ and learn the linear operator, $\mathcal{L}$ which when applied on the measurement vector yields the approximate integral image, with the probabilistic constraint that the error between the approximate integral image and the exact integral image is within allowable limits with high probability. By controlling the allowable error limit, we can obtain measurement matrix of the desired rank. Incorporating a wavelet domain prior model for natural images combined with a relaxation (explained in section 2) allows us to convert the probabilistic constraint into a series of second conic constraints. Rank minimization is a NP-hard problem. Relaxing the objective function to nuclear norm allows to use off-the-shelf convex optimization tools to solve the problem and obtain the measurement matrix and the linear operator.
	    
	    \noindent
		\par{{\bf Contributions:}}
		\textbf{1)}  The main contribution of the paper is a novel framework to recover estimates of integral images from a small number of spatially multiplexed measurements without iterative reconstruction of any auxillary image. We dub the framework {\bf ReFInE} (Reconstruction-Free Integral Image Estimation). \textbf{2)} Leveraging the MGGD (multivariate generalized Gaussian distribution) prior model for the vector of detailed wavelet coefficients of natural images, we propose a nuclear norm minimization formulation to obtain a new specialized measurement matrix. We term the measurements acquired with such a measurement matrix, as {\bf ReFInE} measurements. \textbf{3)} On a large dataset of 4952 images, we present qualitative and  quantitative results to show that high quality estimates of integral images and box-filtered outputs can be recovered from  {\bf ReFInE} measurements in real-time. \textbf{4)}  We show object tracking results, which are comparable to state-of-the-art methods, on a challenging dataset of 50 videos to demonstrate the utility of the box-filtered output estimates in tackling inference problems from SMCs at 1\% measurement rate.
	    
	    \noindent
		\par{{\bf Related Work:}}
		The related previous works in literature follow one of the two themes. Some attempt to tackle inference problems directly from PR measurements without optimizing for the measurement matrix for the inference task at hand, some others attempt to optimize measurement matrix for a particular signal model so as to minimize reconstruction error.
	
		\par
		\textbf{a) Design of measurement matrix:} 
		A closely related work can be found in \cite{duarte2009learning}, wherein a framework is proposed to jointly optimize for a measurement matrix and an overcomplete sparsifying dictionary for small patches of images. Results suggest that better reconstruction results can be obtained using this strategy. However, learning global dictionaries for entire images is not possible, and hence the framework is not scalable. Goldstein \etal \cite{goldstein2013stone} designed measurement matrices called `STOne' Transform which facilitate fast low resolution `previews' just by direct reconstruction, and the hope is that `previews' are of high enough quality so that conventional methods for inference tasks can be applied. Assuming a multi-resolutional signal model for natural images, Chang \etal \cite{chang2009informative} proposed an algorithm to obtain measurements which have the maximum mutual information with natural images. 
		\noindent
		\par
		\textbf{b) Inference problems from CS videos:} A LDS (Linear Dynamical System) based approach was proposed by Sankaranarayanan \etal \cite{CSLDS} to model CS videos and recover the LDS parameters directly from PR measurements. However, the method is sensitive to spatial and view transforms. Calderbank \etal \cite{CLearning} theoretically proved that one can learn classifiers directly from PR measurements, and that with high probability the performance of the linear kernel support vector machine (SVM) classifier operating on the CS measurements is similar to that of the the best linear threshold classifier operating on the original data. A reconstruction-free framework was proposed by Thirumalai \etal \cite{EPFL} to compute optical flow based on correlation estimation between two images, directly from PR measurements. Davenport \etal \cite{Davenport} proposed a measurement domain based correlation filter approach for target classification. Here, the trained filters are first projected onto PR patterns to obtain `smashed filters', and then the PR measurements of the test examples are correlated with these smashed filters. Recently, Kulkarni \etal \cite{kulkarni2015reconstruction} and Lohit \etal \cite{lohit2015reconstruction} extended the `smashed filter' approach to action recognition and face recognition respectively, and demonstrated the feasibility and scalability of tackling difficult inference tasks in computer vision directly from PR measurements.
		  
		\section{Background}
        \noindent
		In this section, we provide a brief background on the probability model for natural images, which we rely on in the paper, and introduce notations required to set up the optimization problem to derive measurement matrix and above referred linear operator.
		 \noindent
		 \par{{\bf Probability Model of natural images:}}
		 There is a rich body of literature which deals with statistical modeling of natural images. We refer to some works which are related to the probability model we use in the paper. Many successful probability models for wavelet coefficients fall under the broad umbrella of Gaussian scale mixtures (GSM) \cite{andrews1974scale}, \cite{wainwright1999scale}. Typically the coefficient space is partitioned into overlapping blocks, and each block is modeled independently as a GSM, which captures the local dependencies. This implicitly gives rise to a global model of the wavelet coefficients. Building on this framework, Lyu \etal \cite{lyu2009modeling} proposed a field of Gaussian scale mixtures (FoGSM) to explicitly model the subbands of wavelet coefficients, while treating each subband independently. However, incorporating such a general model for wavelet coefficient vector makes it very difficult to compute the distribution for even simple functions like a linear function of the wavelet coefficient vector. Therefore, it is vital to assume a prior model which can lead to tractable computation of the distribution. It is well-known that marginal distributions of detailed wavelet coefficients follow generalized Gaussian distribution \cite{mallat1989theory}. We extend this notion to multi-dimensions and model the vector of detailed wavelet coefficients by multivariate generalized Gaussian distribution (MGGD).
		 \noindent
		 \par
		 To put it formally, let ${\bf U}^{T} \in \mathbb{R}^{n \times n}$ be the orthogonal matrix representing the $\log_2(n)$ level wavelet transform, so that $x = {\bf U}w$, where $w \in \mathbb{R}^n$ is the corresponding wavelet coefficient vector. Without loss of generality, we assume that all entries in the first row of ${\bf U}^{T}$ are $1/\sqrt{n}$ so that the first entry in $w$ corresponds to $\sqrt{n}$ times the mean of all entries in $x$. Also we denote the rest $n-1$ rows in ${\bf U}^{T}$ by ${\bf U}_{2:n}^{T}$. Now we can write $w = [\sqrt{n}\bar{x}, w_d]$, where $\bar{x}$ is the mean of $x$ and $w_d$ is the vector of detailed coefficients. As explained above, the probability distribution of $w_d$ (MGGD) is given by 
		 \begin{equation}
		 \label{MGGD}
		 f(w) = K|{\bf \Sigma}_{w_d}|^{-0.5}exp(-(w_d^{T}{\bf \Sigma}_{w_d}^{-1} w_d)^\beta),
		 \end{equation}
		 where ${\bf \Sigma}_{w_d}$, commonly known as the scatter matrix, is equal to rank(${\bf \Sigma}_{w_d}$) $\Gamma{((2+n-2)/2\beta)}/\Gamma{((2+n)/2\beta)}$ times the covariance matrix of $w_d$, $\beta \in (0,1]$, and $K$ is a normalizing constant. For $\beta = 1$, we obtain the probability distribution for the well-known multivariate Gaussian distribution. In the following we briefly provide a background regarding the multivariate generalized Gaussian distribution.
		 
		 \par A linear transformation of the multivariate generalized Gaussian random vector is also a multivariate generalized Gaussian random vector.
		 \begin{proposition}\cite{frahm2004generalized} 
		 	Let $u$ be the a $n \times 1$ multivariate generalized Gaussian random vector with mean $\mu_u \in \mathbb{R}^n$, and scatter matrix, ${\bf \Sigma}_{u} \in \mathbb{R}^{n \times n}$. Let $A$ be a $l \times n$ full rank matrix. Then the $l \times 1$ random vector, $v = Au$ has the multivariate generalized Gaussian distribution with mean $\mu_v = A\mu_u \in \mathbb{R}^l$, and scatter matrix, ${\bf \Sigma}_{v} = A{\bf \Sigma}_{u}A^T \in \mathbb{R}^{l \times l}$. 	
		 \end{proposition}
		 If $v$ is a univariate generalized Gaussian random variable, then the probability of $v$ falling in the range of $[\delta-\mu_v, \delta+\mu_v]$, for $\delta \geq 0$, can be found in terms of lower incomplete gamma function.
		 \begin{proposition}
		 	If $v$ is a univariate generalized Gaussian random variable with mean $\mu_v \in \mathbb{R}$ and scatter matrix, ${\bf \Sigma}_{v} \in \mathbb{R}$, then the probability of $v$ falling in the range of $[-\delta+\mu_v, \delta+\mu_v]$, for $\delta \geq 0$, is given by,
		 	\begin{equation}
		 	\mathbb{P}(|v - \mu_v| \leq \delta) = 2\gamma\left(\frac{1}{2\beta}, \left(\frac{\delta}{{\bf \Sigma}_{v}} \sqrt{\frac{\Gamma(\frac{3}{2\beta})}{\Gamma(\frac{1}{2\beta})}} \right)^{2\beta} \right),
		 	\end{equation} 
		 \end{proposition} 
		 where $\gamma(.,.)$ is the lower incomplete gamma function, and $\Gamma(.)$ is the ordinary gamma function.

        \noindent	
		\par{{\bf Preliminaries:}}
		Let ${\bf H} \in \mathbb{R}^{n \times n}$  be the block Toeplitz matrix representing the integral operation so that the integral image, $I = {\bf H}x  \in \mathbb{R}^{n}$, and $h_i^{T}$ for $i = 1,..,n$ be the rows of ${\bf H}$. Hence, $I_i$ equals $h_i^{T}x$. We wish to recover the approximate integral image, $\hat{I}$ from the measured vector $y = \phi x$, just by applying a linear operator $\mathcal{L}$ on $y$, so that $\hat{I} = \mathcal{L}y$. For reasons which will be apparent soon, we assume $\mathcal{L} = {\bf H}(\phi^d)^{T}$, where $\phi^d \in \mathbb{R}^{m \times n} $ such that rank$(\phi^d)$ = rank$(\phi)$. We call $\phi^d$ as the dual of $\phi$. Thus by construction $\mathcal{L} \in \mathbb{R}^{n \times m}$. The value at location $i$ in the approximate integral image is given by $\hat{I}_i = h_i^{T}(\phi^d)^{T}\phi x$. The distortion in integral image at location $i$ is given by $d_i = \hat{I}_i - I_i = h_i^{T}((\phi^d)^{T}\phi x - x)$. Noting ${\bf Q} = (\phi^d)^{T}\phi$, and $n \times n$ identity matrix by ${\bf I}$, the distortions can be compactly written as $d_i = h_i^{T}({\bf Q}-{\bf I})x$. We call ${\bf d} = [d_1,...,d_n]$ as distortion vector.

		\section{Optimization problem}
		Our aim is to search for a measurement matrix $\phi$ of minimum rank such that distortions, $d_i$ are within allowable limits for all $i$, jointly, with high probability. ${\bf Q}$ by construction is the product of two matrices of identical ranks, $\phi$ and $(\phi^d)^T$. Hence, we have the relation, rank(${\bf Q}$) = rank($\phi$). Inspired by the phase-lifting technique used in \cite{candes2013phaselift} and \cite{hegdenumax}, instead of minimizing the rank of $\phi$, we minimize the rank of ${\bf Q}$. Now, we can formally state the optimization problem as follows.

		\begin{equation}\label{opti2}
		\begin{split}
		\underset{{\bf Q}}{\text{minimize}} \quad \mathrm{rank}({\bf Q}) \quad \quad \quad \quad \quad \\
		\text{s.t} \quad \mathbb{P} (|d_1| \le \delta_1,..,|d_i| \leq \delta_i.., |d_n| \leq \delta_n ) \geq 1-\epsilon,
		\end{split}
		\end{equation}
		where $\delta_i \ge 0$ denotes the allowable limit of distortion at  location $i$ of integral image, and $0 < \epsilon < 1$. Once ${\bf Q}^{*}$ is found, we show later in the section that the SVD decomposition of ${\bf Q}^{*}$ allows us to write ${\bf Q}^{*}$ as a product of two matrices of identical ranks, thus yielding both the measurement matrix, $\phi^{*}$ and the desired linear operator, $\mathcal{L}^{*}$. The constraint in (\ref{opti2}) is a probabilistic one. Hence to compute it, one needs to assume a statistical prior model for $x$. Using the model in \ref{MGGD}, and its properties given in proposition (1) and (2), we arrive at a solvable optimization problem.

	    \noindent
		\par{{\bf Computation of probabilistic constraint in (\ref{opti2}):}}
		Substituting for $x$, we can write the distortion at location $i$ as $d_i = h_i^{T}({\bf Q}-{\bf I}){\bf U}w$. We let all the entries in the first row of $\phi$ and $\phi^d$ to be equal to $1/\sqrt{n}$, so that one of the $m$  measurements is exactly equal to $\sqrt{n}\bar{x}$. Further, we denote the rest $m-1$ rows of the two matrices by $\phi_{2:m}$ and $\phi^d_{2:m}$. Now, if we restrict $\phi_{2:m}$ and $\phi^{d}_{2:m}$ to be respectively equal to ${\bf C} {\bf U}_{2:n}^{T}$ and  ${\bf D} {\bf U}_{2:n}^{T}$ for some ${\bf C}$, ${\bf D}$ in $\mathbb{R}^{m-1 \times n-1}$, then from basic linear algebra we can show that $d_i = h_i^{T}({\bf P}-{\bf I}){\bf U}_{2:n}^{T}w_d$,
		where ${\bf P} = (\phi^d_{2:m})^T\phi_{2:m}$. It is easy to see that ${\bf Q} = {\bf P} + \frac{1}{n}{\bf O}$, and rank(${\bf Q}$) = rank(${\bf P}$) + 1, where ${\bf O}$ is the matrix with all its entries equal to unity (see Appendix A for details). Hence we can replace the objective function in (\ref{opti2}) by rank($\bf P$). Rank minimization is a non-convex problem. Hence we relax the objective function to nuclear norm, as is done typically. To compute the constraint in (\ref{opti2}), one needs to first compute the joint probability of ${\bf d} = [d_1,..,d_n]$, and then compute a $n$ dimensional definite integral. Now that $d_i$'s are linear combinations of $w_d$, it follows from proposition 1, that ${\bf d}$ also has a multivariate generalized Gaussian distribution. However, no closed form for the definite integral is known. Hence, we relax the constraint by decoupling it into $n$ independent constraints, each enforcing the constraint that the distortion at a specific location is to be within allowable limits with high probability, independent of the distortions at other locations. The optimization with relaxed constraints is thus given by 
	
		\begin{equation}\label{opti3}
		\begin{split}
		\underset{{\bf P}}{\text{minimize}} \quad
		\norm{\bf{P}}_* \quad \quad \quad \quad \\
		\text{s.t} \quad
		\mathbb{P} (|d_i| \le \delta_i) \geq 1-\epsilon, \quad  i = 1,..,n.
		\end{split}
		\end{equation} 
		Now, $d_i = h_i^{T}({\bf P}-{\bf I}){\bf U}_{2:n}^{T}w_d$, is a linear combination of the entries of $w_d$. From the proposition 2, $d_i$ has a one-dimensional generalized Gaussian distribution with zero mean and scatter parameter, $\norm{{\bf \Sigma}^{1/2}_{w_d}{\bf U}_{2:n}^T  ({\bf P}-{\bf I})^T h_i}$, and the probability in equation \ref{opti3} can be explicitly written as follows.
		\begin{equation}\label{probconst}
		\scriptsize{\begin{split}
			\mathbb{P} (|d_i| \le \delta_i) \quad \quad \quad \quad \quad \quad \quad \quad \quad \quad \quad \quad \quad \quad \quad \quad \quad \quad \quad \quad \quad \quad \\
			= 2 \gamma \left(\frac{1}{2\beta},\left(\frac{\delta_i}{\norm{{\bf \Sigma}^{1/2}_{w_d}{\bf U}_{2:n}^T{\bf P}^T h_i-{\bf \Sigma}^{1/2}_{w_d}{\bf U}_{2:n}^T h_i}_2}\sqrt{\frac{\Gamma(\frac{3}{2\beta})}{\Gamma(\frac{1}{2\beta})}} \right)^{2\beta}\right).
			\end{split}}
		\end{equation}
	The optimization problem now can be rewritten as 
		\begin{align*}\label{opti4_con}
	\underset{{\bf P}}{\text{minimize}}
			\norm{\bf{P}}_* \quad   \quad \text{s.t}  \quad \quad \quad \quad \quad \quad \quad 
			\\
		{\scriptsize 2 \gamma \left(\frac{1}{2\beta},\left(\frac{\delta_i}{\norm{{\bf \Sigma}^{1/2}_{w_d}{\bf U}_{2:n}^T{\bf P}^T h_i-{\bf \Sigma}^{1/2}_{w_d}{\bf U}_{2:n}^T h_i}_2}\sqrt{\frac{\Gamma(\frac{3}{2\beta})}{\Gamma(\frac{1}{2\beta})}} \right)^{2\beta}\right)} \\ \geq 1-\epsilon  \quad \quad \quad \quad \quad \quad \quad \quad \quad \quad \quad \quad \quad \quad.
	    \end{align*} We compactly write ${\bf \Sigma}^{1/2}_{w_d}{\bf U}_{2:n}^T{\bf P}^T h_i$ as $\mathcal{A}_i({\bf P})$, ${\bf \Sigma}^{1/2}_{w_d}{\bf U}_{2:n}^T h_i$ as $b_i$  and $\frac{\delta_i}{(\gamma^{-1}(\frac{1}{2\beta},\frac{1-\epsilon}{2}))^{\frac{1}{2\beta}}} \sqrt{\frac{\Gamma(\frac{3}{2\beta})}{\Gamma(\frac{1}{2\beta})}}$ as $\Delta_i$. Plugging above in equation (\ref{probconst}), and rearranging terms, we have 
		\begin{equation}\label{opti4}
		\begin{split}
		\underset{{\bf P}}{\text{minimize}} \quad
		\norm{\bf{P}}_* \quad \quad \quad \quad \\
		\text{s.t} \quad
		\norm{\mathcal{A}_i({\bf P})-b_i}_2 \le \Delta_i \quad i = 1,..,n. \\
		\end{split}
		\end{equation}
		We can rewrite the problem in conic form as below.

		\begin{equation}\label{opti5}
		\begin{split}
		(P1) \quad \underset{\bf{P}}{\text{minimize}} \quad
		\norm{\bf{P}}_* \quad \quad \text{s.t} \\
	    \left[
		\begin{array}{lr}
		b_i-\mathcal{A}_i(\bf{P}) \\
		\Delta_i 
		\end{array}
		\right] \in \mathcal{K}_i, \quad  i = 1,..,n,
         \end{split}
		\end{equation} 
		where $\mathcal{K}_i$ is a second order cone $\mathcal{K}_i = \{(x_i,t_i) \in \mathbb{R}^{n+1} : \norm{x_i} \le t_i \}$. Let ${\bf b} = [b_1,..,b_n]$, and ${\mathcal{A}({\bf P}) = [\mathcal{A}_1({\bf P}),..,\mathcal{A}_n({\bf P})]}$. Let $\mathcal{A}^*$  denote the adjoint of the linear operator $\mathcal{A}$.
		It is easy to recognize that the optimization above is a convex problem, since nuclear norm is convex and the constraints enforce finite bounds on the norms of affine functions of the decision variable, ${\bf {P}}$ and hence are also convex.
		Even though the constraints are second-order cone constraints, the standard second order conic programming methods cannot be used to solve $(P1)$ since nuclear norm is non-smooth. The nuclear norm is smoothened by the addition of a square of Forbenius norm of the matrix, and is replaced by $\tau\norm{\bf{P}}_* + \frac{1}{2}\norm{\bf{P}}_F^2$, where $\tau > 0$. The optimization problem with the smoothened objective function is given in \ref{opti6}. 
		\begin{equation}\label{opti6}
		\begin{split}
		(P2) \quad
		\underset{\bf{P}}{\text{minimize}} \quad
		\tau\norm{\bf{P}}_* + \frac{1}{2}\norm{\bf{P}}_F^2 \quad \quad \quad \\
		\text{s.t}
		\left[
		\begin{array}{lr}
		b_i-\mathcal{A}_i(\bf{P}) \\
		\Delta_i 
		\end{array}
		\right] \in \mathcal{K}_i, \quad \quad  i = 1,..,n.
		\end{split}
		\end{equation} 
		Recently, many algorithms \cite{liu2012implementable,cai2010singular} have been developed to tackle nuclear norm minimization problem of this form in the context of matrix completion. We use SVT (singular value thresholding)  algorithm \cite{cai2010singular} to solve $(P2)$.
    \par{{\bf SVT iteration to solve ($P2$):}}
    Here, we first briefly describe the SVT algorithm for smoothened nuclear norm minimization with general convex constraints, and later we show how we adapt the same to our problem $P2$ which has $n$ second-order constraints. Let the smoothened nuclear norm with general convex constraints, be given as below.
    	\begin{equation}\label{opti7}
    	\begin{split}
 \underset{\bf{P}}{\text{minimize}} \quad
    	\tau\norm{\bf{P}}_* + \frac{1}{2}\norm{\bf{P}}_F^2 \quad \quad \quad \quad \\
    	\text{s.t}  \quad
    	 f_i({\bf P}) \le 0, \quad  i = 1,..,n,
    	\end{split}
    	\end{equation} 
where $f_i({\bf P})\le0$,  $i=1,..,n$ denote the $n$ convex constraints. Let $\mathcal{F}({\bf P}) = [f_1({\bf P}),..,f_n({{\bf P}})]$. The SVT algorithm for the \ref{opti7} with the modified objective function is given as below.
 \begin{equation}\label{uzawa1}
 \begin{rcases}
 \begin{array}{lr}
 \begin{array}{lr}
 {\bf P}^k = \underset{\bf{P}} \argmin \quad  {\tau\norm{\bf{P}}_* + \frac{1}{2}\norm{\bf{P}}_F^2 + \langle {\bf z}^{k-1}, \mathcal{F}({\bf P}) \rangle}
 \end{array} \\
 \begin{array}{lr}
 {\bf z_i}^k
 \end{array}
 = P_i\left({\bf z_i}^{k-1}  + \eta^{k}f_i({\bf P}^k) \right), \quad  i = 1,..,n
 \end{array}
 \end{rcases}
 \end{equation}
where ${\bf z}^k$ is a short form for $[{\bf z_1}^k,..,{\bf z_n}^k]$, and $P_i({\bf q})$ denotes the projection of ${\bf q}$ onto the convex set defined by the constraint $f_i({\bf P}) \le 0$. Let ${\bf z_i}^k = [{\bf y_i}^k, s_i^k]$, so that the vector ${\bf y_i}^k$ denotes the first $n$ elements of ${\bf z_i}^k$ and $s_i^k$ denotes the last element of ${\bf z_i}^k$. Let ${\bf y}^k$ be a short form for $[{\bf y_1}^k,..,{\bf y_n}^k]$, and $s^k$ is a short form for $[s_1^k,..,s_n^k]$. To obtain a  explicit form of the update equations in \ref{uzawa1} for our problem, $P1$, let us consider the first equation of the same. $\mathcal{F}(\bf{P})$ for $P1$ is given by $[b_1 - \mathcal{A}_1({\bf P}), \Delta_1,..,b_n - \mathcal{A}_n({\bf P}),\Delta_n]^T$. We substitute for $\mathcal{F}({\bf P})$, and after removal of the terms not involving ${\bf P}$, we have
\begin{equation}\label{uzawa2}
{\bf P}^k = \underset{\bf{P}} \argmin \quad  {\tau\norm{\bf{P}}_* + \frac{1}{2}\norm{\bf{P}}_F^2 + \langle {\bf y}^{k-1}, {\bf b}- \mathcal{A}({\bf P}) \rangle}.
\end{equation}
Equation \ref{uzawa2} can be rewritten as below.
\begin{align*}\label{uzawa2}
{\bf P}^k = \underset{\bf{P}} \argmin \quad  \tau\norm{{\bf P}}_* + \frac{1}{2}\norm{{\bf P} - \mathcal{A}^*({\bf y}^{k-1})}_F^2 \\ - \frac{1}{2}\norm{\mathcal{A}^*({\bf y}^{k-1})}_F^2 + \langle{\bf P}, \mathcal{A}^*({\bf y}^{k-1}) \rangle \\ + \langle {\bf y}^{k-1}, {\bf b} \rangle - \langle{\bf y}^{k-1}, \mathcal{A}({\bf P}) \rangle.
\end{align*}
Removing the terms not involving ${\bf P}$ and noting that $\langle{\bf P}, \mathcal{A}^*({\bf y}^{k-1}) \rangle = \langle{\bf y}^{k-1}, \mathcal{A}({\bf P}) \rangle$, we have the following.
\begin{equation}\label{uzawa3}
{\bf P}^k = \underset{\bf{P}} \argmin \quad  \tau\norm{\bf{P}}_* +  \frac{1}{2}\norm{{\bf P} - \mathcal{A}^*({\bf y}^{k-1})}_F^2.
\end{equation}    
Before we write down the solution to equation \ref{uzawa3}, we first define  $\mathcal{D}_\tau$, the singular value shrinkage operator.  Consider the SVD of a matrix ${\bf X}$, given by ${\bf X} = {\bf W}{\bf \Sigma V}^T$. Then for $\tau \ge 0$, the singular value shrinkage operator, $\mathcal{D}_\tau$ is given by $\mathcal{D}_\tau({\bf X}) = {\bf W}\mathcal{D}_\tau({\bf \Sigma}){\bf V}^T,
\mathcal{D}_\tau({\bf \Sigma}) = diag(\{(\sigma_i-\tau)_+\})$,
where $t_+$ = max(0,$t$).  The solution to equation \ref{uzawa3} is given by ${\bf P}^k = \mathcal{D}_\tau(\mathcal{A}^*({\bf y}^{k-1}))$. Now, it remains to calculate $\mathcal{A}^*({\bf y}^{k-1})$. We achieve it according to the following. Consider $\langle \mathcal{A}^*({\bf y}^{k-1}), {\bf P} \rangle$.
\begin{align*}
 \langle  {\bf P}, \mathcal{A}^*({\bf y}^{k-1}) \rangle \\
= \langle \mathcal{A}({\bf P}), {\bf y}^{k-1} \rangle
=  \sum_{i=1}^{n} \langle \mathcal{A}_i({\bf P}), {\bf y_i}^{k-1} \rangle \\
=   \sum_{i=1}^{n}\langle {\bf P} , \mathcal{A}_i^*({ {\bf y_i}^{k-1}}) \rangle 
= \langle {\bf P}, \sum_{i=1}^{n}\mathcal{A}_i^*({\bf y_i}^{k-1}) \rangle
\end{align*}
Hence, we have $\mathcal{A}^*({\bf y}^{k-1}) = \sum_{i=1}^{n}\mathcal{A}_i^*({\bf y_i}^{k-1})$. Thus, the first equation of SVT iteration for our problem is given by 
\begin{equation}
{\bf P}^k = \mathcal{D}_\tau\left(\sum_{i=1}^{n}\mathcal{A}_i^*({\bf y_i}^{k-1})\right).
\end{equation}
Using basic linear algebra,  it can be shown that $\mathcal{A}_i^*({\bf y_i}^{k-1}) = {\bf U}_{2:n}(\Sigma_{w_d}^{1/2})^T {\bf y_i}^{k-1} h_i^T$ (see Appendix B for details).
\par We now provide the projection onto the convex cone $\mathcal{K}_i$. The projection operator, $P_{\mathcal{K}_i}$ as derived in \cite{fukushima2002smoothing} is given as follows.
		
		\begin{equation}
		P_{\mathcal{K}_i} : (x,t) \mapsto 
		\begin{array}{lr}
		(x,t),  \quad \quad \quad \quad \quad \quad \norm{x} \le t, \\
		\frac{\norm{x}+t}{2\norm{x}} (x,\norm{x}), \quad -\norm{x} \le t \le \norm{x},   \\
		(0,0), \quad \quad \quad \quad \quad \quad    t \le -\norm{x}.  
		\end{array}
		\end{equation}
		
		To solve $(P2)$, starting with
		$\left[
		\begin{array}{lr}
		{\bf y}^0_i \\
		s^0_i
		\end{array}
		\right] = {\bf 0}$ for all $i = 1,..n$, the $k^{th}$ SVT iteration is given by (\ref{uzawa}).
		\begin{equation}\label{uzawa}
		\begin{rcases}
		\begin{array}{lr}
		\begin{array}{lr}
		{\bf P}^k = \mathcal{D}_\tau\left(\sum\limits_{i=1}^{n}{\bf U}_{2:n}(\Sigma_{w_d}^{1/2})^T {\bf y_i}^{k-1} h_i^T\right) 
		\end{array} \\
		\left[
		\begin{array}{lr}
		{\bf y}^k_i \\
		s^k_i
		\end{array}
		\right] = P_{\mathcal{K}_i}\left(\left[\begin{array}{lr}
		{\bf y}^{k-1}_i \\
		s^{k-1}_i
		\end{array}\right] + \eta^k \left[
		\begin{array}{lr}
		b_i-\mathcal{A}_i({\bf P}^k) \\
		-\Delta_i 
		\end{array}
		\right]    \right),
		\end{array}
		\end{rcases}
		\end{equation}
		where, $\mathcal{A}^*_i$ are the adjoints of linear operators $\mathcal{A}_i$.
		For the iterations (\ref{uzawa}) to converge, we need to choose the step sizes, $\eta^k \le \frac{2}{\norm{\mathcal{A}}_2^2}$, where $\norm{\mathcal{A}}_2$ is the spectral norm of the linear transformation $\mathcal{A}$ \cite{cai2010singular}.
		
		Once the solution ${\bf P}^*$ is found, using the relation noted earlier in the section, we have ${\bf Q}^* = {\bf P}^* + \frac{1}{n} {\bf O}$. Having obtained ${\bf Q}^*$, the task now is to express it into a product of two matrices of identical ranks. This is done almost trivially as follows. Noting the singular value decomposition of ${\bf Q^*}$ as ${\bf Q^*} = {\bf W}_{M}{\bf \Sigma}_{M}{\bf V}^{T}_{M}$, where ${\bf \Sigma}_{M} =  diag\{\lambda_1,..,\lambda_M\}$ denote the diagonal matrix with the non-zero singular values arranged along its diagonal, and ${\bf W}_{M}$ and ${\bf V}^{T}_{M}$ are the matrices whose columns are the left and right singular vectors respectively. We can choose $\phi^* = {\bf \Sigma}_{M}^{1/2} {\bf V}^{T}_{M}$, so that $((\phi^d)^{*})^{T} = {\bf W}_{M} {\bf \Sigma}^{1/2}_{M}$ and rank($\phi^*$) = rank($(\phi^d)^*$) = rank(${\bf Q}^*$). We, henceforth refer to $\phi^{*}$ as {\bf ReFInE} $\phi$, and the corresponding measurements, $y = \phi^{*}x$ as {\bf ReFInE} measurements.  The desired linear operator $\mathcal{L}^{*}$ is given by ${\bf H}{\bf W}_{M} {\bf \Sigma}^{1/2}_{M}$. By construction, the approximate integral image $\hat{I}$ is given by $\hat{I} = \mathcal{L}^{*}y = ({\bf H}{\bf W}_{M} {\bf \Sigma}^{1/2}_{M})y$.
		
		\par{{\bf Computational Complexity:}}
		Since the length of the image $x$ is $n$, the number of entries in ${\bf P}$ is $n^2$. Hence, the dimension of the optimization problem $P2$ is $n^2$. This means that if we are to optimize for a measurement matrix to operate on an image of size, $256 \times 256$, the dimension of the optimization problem would be $2^{32}$! Optimizing over a large number of variables is computationally expensive, if not impractical. Hence we propose the following suboptimal solution to obtain the measurement matrix. We divide the image into non-overlapping blocks of fixed size, and sense each block using a measurement matrix, optimized for this fixed size, independently of other blocks. Let the image, $x$ be divided into $B$ blocks, $x_1, x_2, .. ,x_B$, each of a fixed size, $f \times f$, and $\phi_f \in \mathcal{R}^{m \times f^2}$ be the measurement matrix optimized for images of size $f \times f$, and $(\phi^d_f)^T$ be the corresponding dual matrix. Then the `{\bf ReFInE}' measurements, $y$ are given by the following,
    \begin{equation}
	y = \begin{bmatrix}
	y_1  \\
	y_2  \\
	\vdots  \\
	y_B
	\end{bmatrix}  =
	\begin{bmatrix}
 \phi_{f} & {\bf 0} & \cdots & {\bf 0} \\
	{\bf 0} &  \phi_{f}  & \cdots & {\bf 0} \\
		\vdots  & \vdots  & \ddots & \vdots  \\
	{\bf 0} & {\bf 0} & \cdots &  \phi_{f} 
	\end{bmatrix} 	\begin{bmatrix}
	x_1  \\
	x_2  \\
	\vdots  \\
    x_B
	\end{bmatrix}. 
	\end{equation}    
Once the measurements, $y$ are obtained, the integral image, $\hat{I}$ is given by 
\begin{equation}
\hat{I} = {\bf H} \begin{bmatrix}
(\phi^d_{f})^T & {\bf 0} & \cdots & {\bf 0} \\
{\bf 0} &  (\phi^d_{f})^T & \cdots & {\bf 0} \\
\vdots  & \vdots  & \ddots & \vdots  \\
{\bf 0} & {\bf 0} & \cdots &  (\phi^d_{f})^T
\end{bmatrix} 	\begin{bmatrix}
y_1  \\
y_2  \\
\vdots  \\
y_B
\end{bmatrix}. 
\end{equation} 
		
		\section{Experiments}

		Before we can conduct experiments to evaluate our framework, we first need to estimate the parameters of the probability model in \ref{MGGD}.
		Estimating parameters of the probability model and optimizing measurement matrices for any arbitrary large sized image blocks is not practical since the former task requires an enormous amount of image data and the latter requires prohibitive amount of memory and computational resources. Hence, we fix the block size to be $32 \times 32$ images. The scatter matrix ${\bf \Sigma}_{w_d}$ is a scalar multiple of the covariance matrix of $w_d$. Hence it suffices to compute the covariance matrix. To this end, we first downsample all the 5011 training images in PASCAL VOC 2007 dataset \cite{pascal-voc-2007} to a size of $32 \times 32$, so that $n$ = 1024 and then obtain the level 7 Daubechies wavelet coefficient vectors. We compute the sample covariance matrix of thus obtained wavelet coefficient data. For various values of $\beta$, we evaluate the $\chi^2$ distance between the histograms of the individual wavelet coefficients and their respective theoretical marginal distributions with the variances computed above. We found for $\beta = 0.68$, the distance computed above is minimum.

		\noindent
		\par{{\bf Computing measurement matrix:}} 
		To obtain a measurement matrix, we need to input a desired distortion vector $\delta$ to the optimization problem in $(P2)$. The desired distortion vector is computed according to the following. We first perform principal component analysis (PCA) on the downsampled 5011 training images in the PASCAL VOC 2007 dataset \cite{pascal-voc-2007}. We use only the top 10 PCA components as $\phi$ to `sense' these images. We obtain the desired distortion vector by first assuming $\phi^{d} = \phi$ and calculating distortions, $|d^j_i|$ at each location for all training images, $j =1,..,5011$. Now, the entry in location $i$ of the desired ${\bf \delta}$ is given by the minimum value $\alpha$, so that $95\%$ of the values, $|d^j_i|, j =1,..,5011$ are less than $\alpha$.  We use $\epsilon =0.95$ and solve $(P2)$ to obtain ${\bf P}^*$, and hence also ${\bf Q}^*$. The rank of {\bf ReFInE} $\phi$ is simply the rank of ${\bf Q}^*$. 
		
			
		
		\noindent
		\par
\begin{figure*}
			\centering
			\subfigure[]{\includegraphics[height=6em,width=8em]{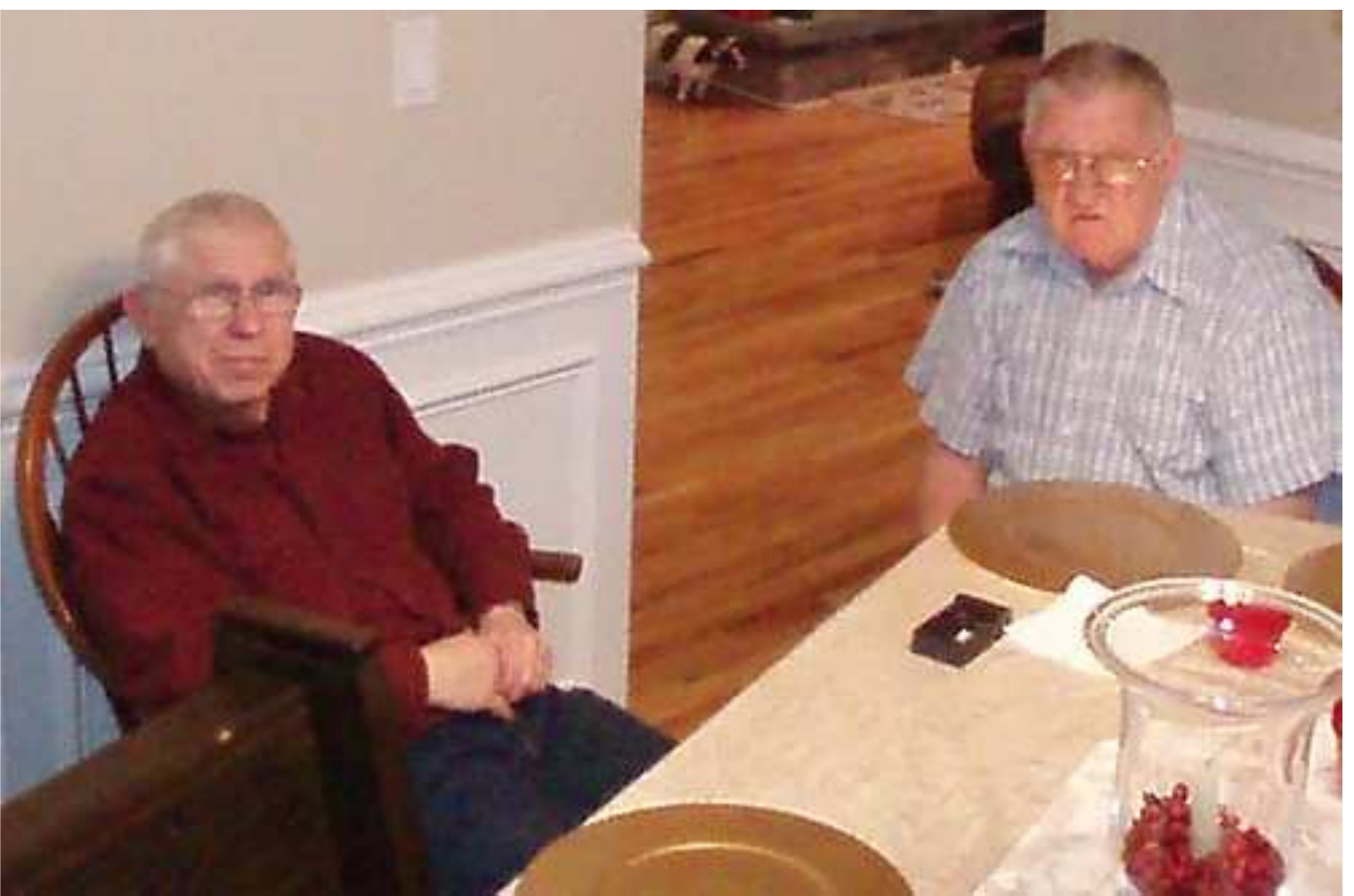}
				\label{fig:twomen}
			}
			\subfigure[]{\includegraphics[height=6em,width=8em]{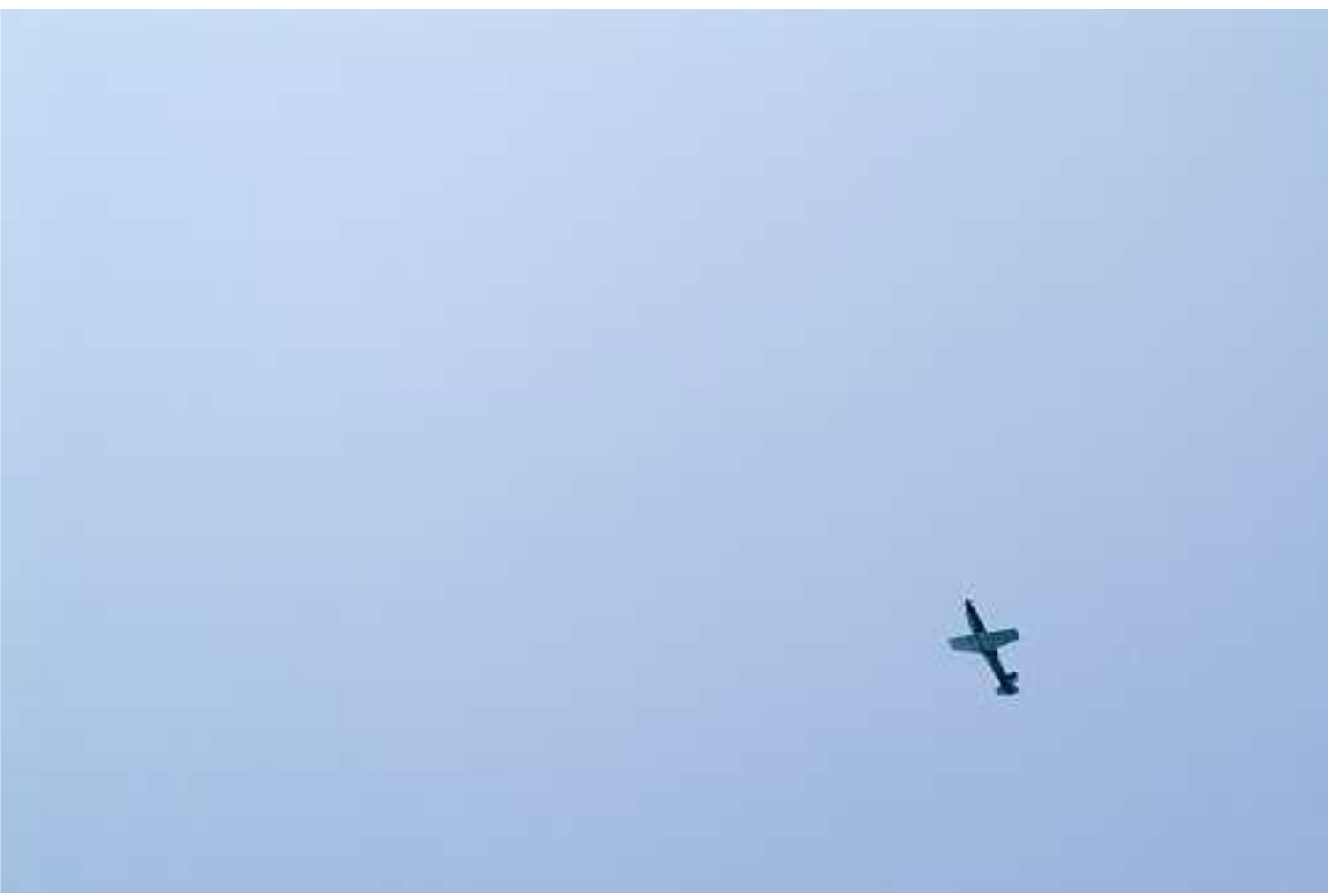}
				\label{fig:plane}
			}
			\subfigure[]{\includegraphics[height=6em,width=8em]{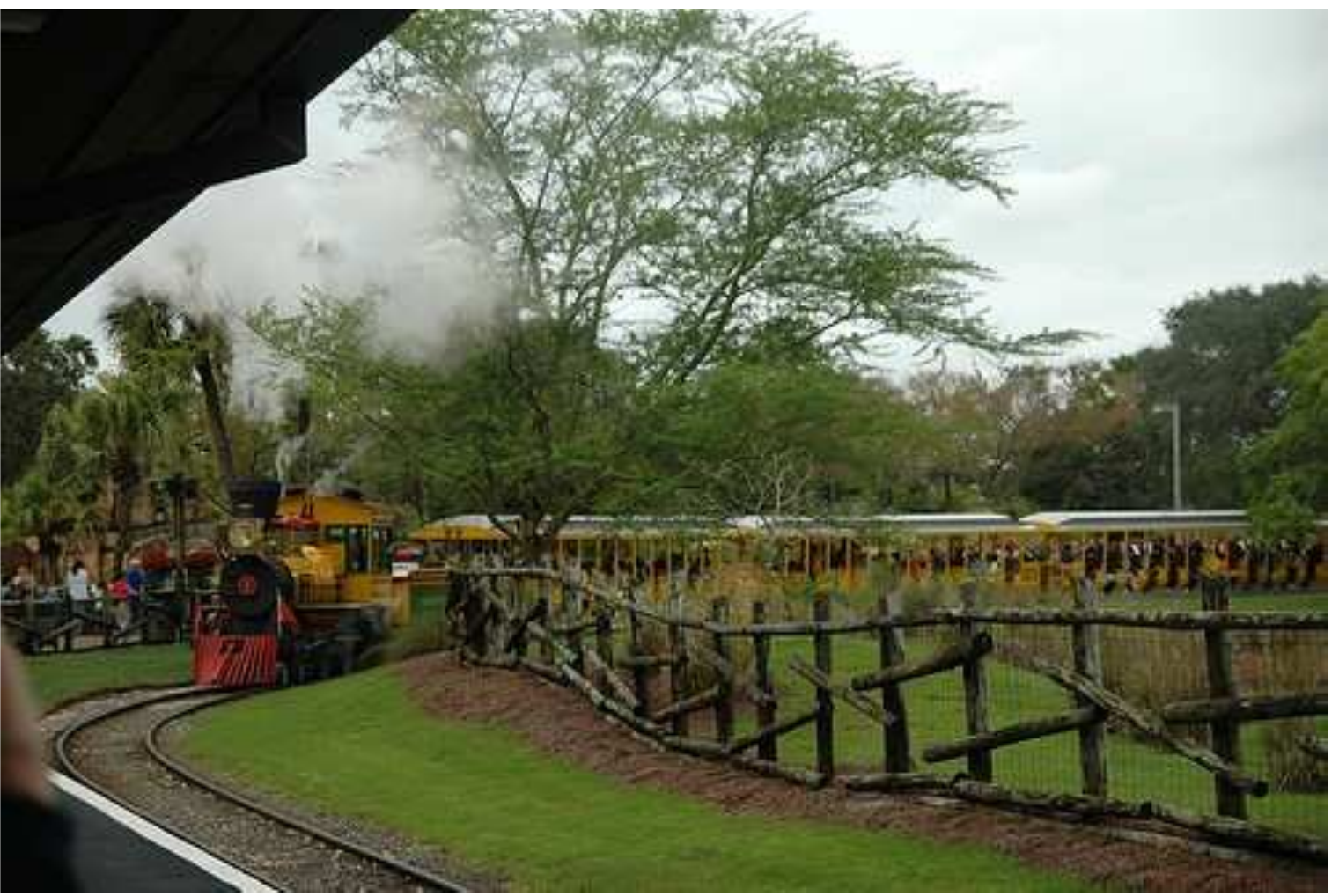}
				\label{fig:train}
			}
			\subfigure[]{\includegraphics[height=6em,width=8em]{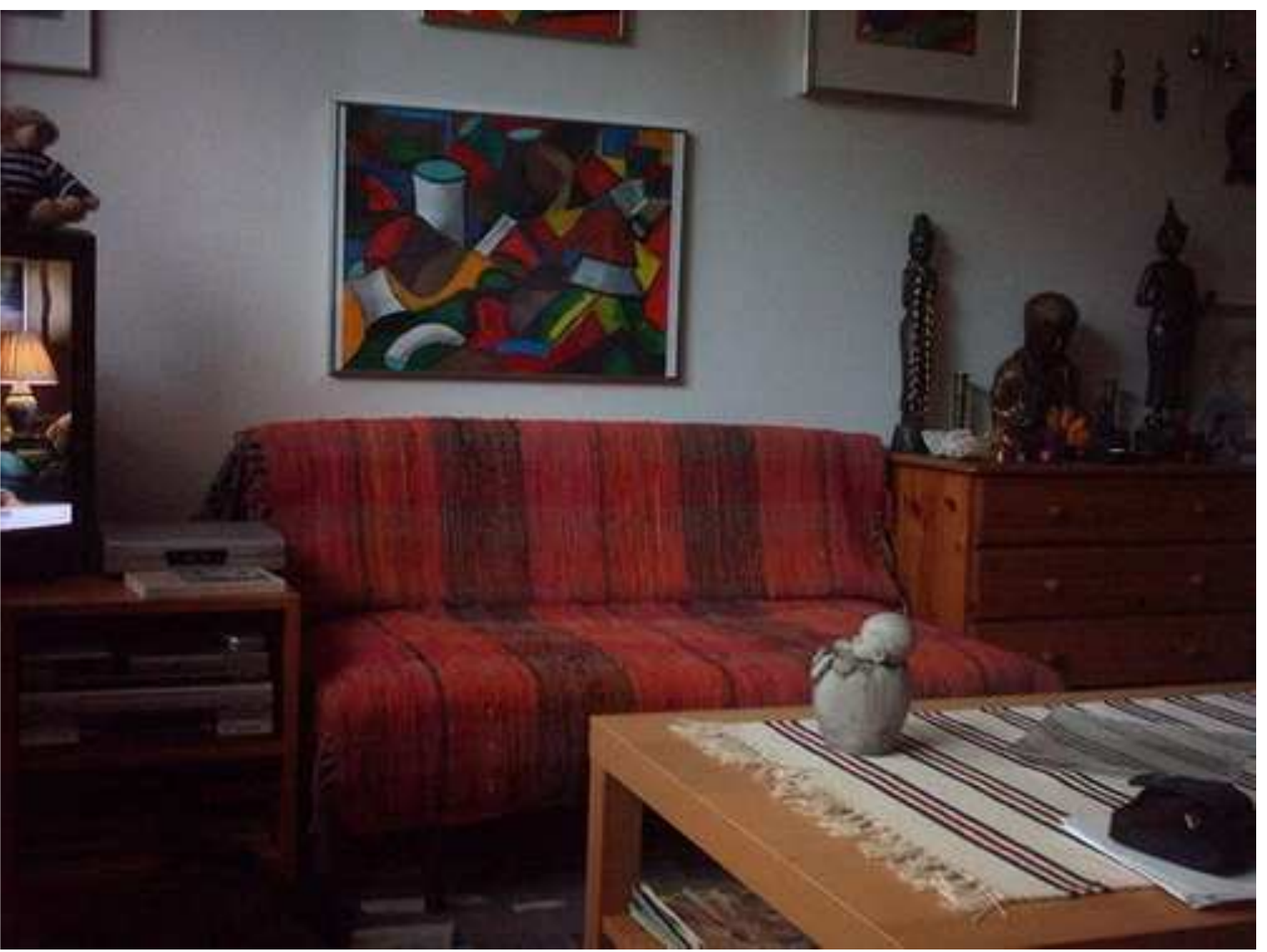}
				\label{fig:room}
			}
			\vspace{0.5cm}
			\caption{(Four images (L-R: `Two Men', `Plane', `Train' and `Room') are randomly chosen for presenting qualitative and quantitative results.}
		\end{figure*}

\begin{figure*}
            \centering
			\subfigure[]{\includegraphics[height=14em,width=20em]{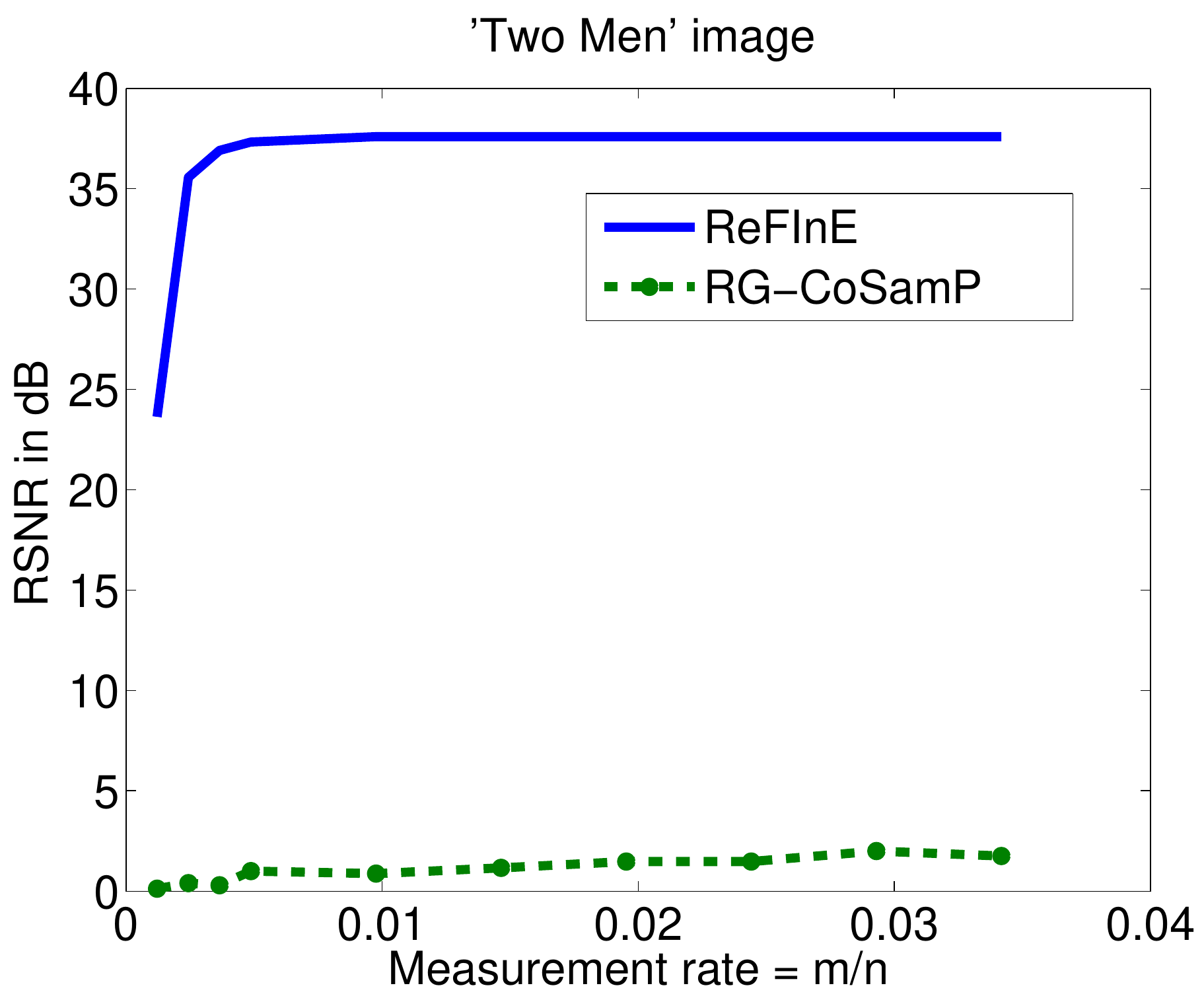}
				\label{fig:subfig11}
			}
			\subfigure[]{\includegraphics[height=14em,width=20em]{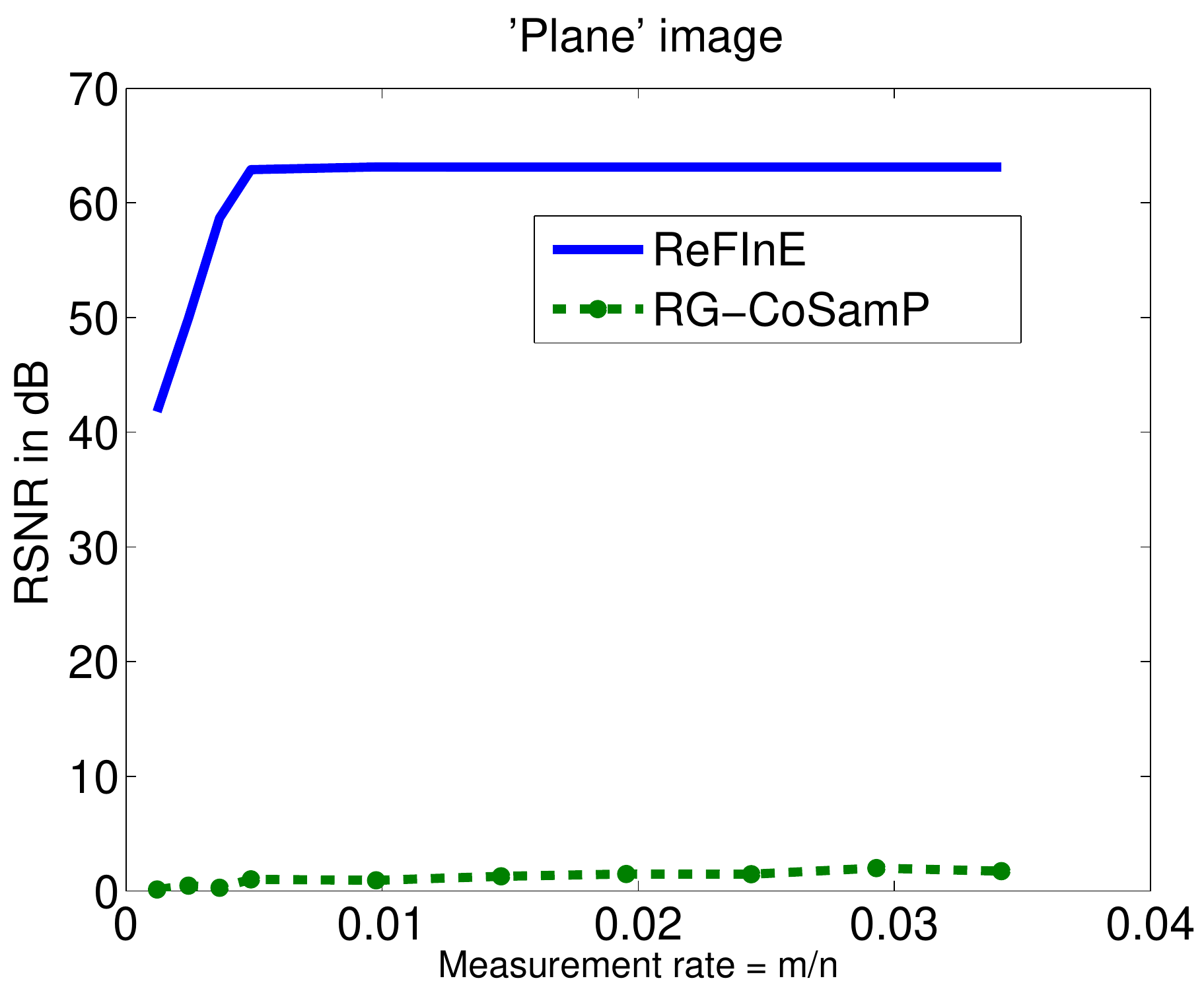}
				\label{fig:subfig12}
			}
			\subfigure[]{\includegraphics[height=14em,width=20em]{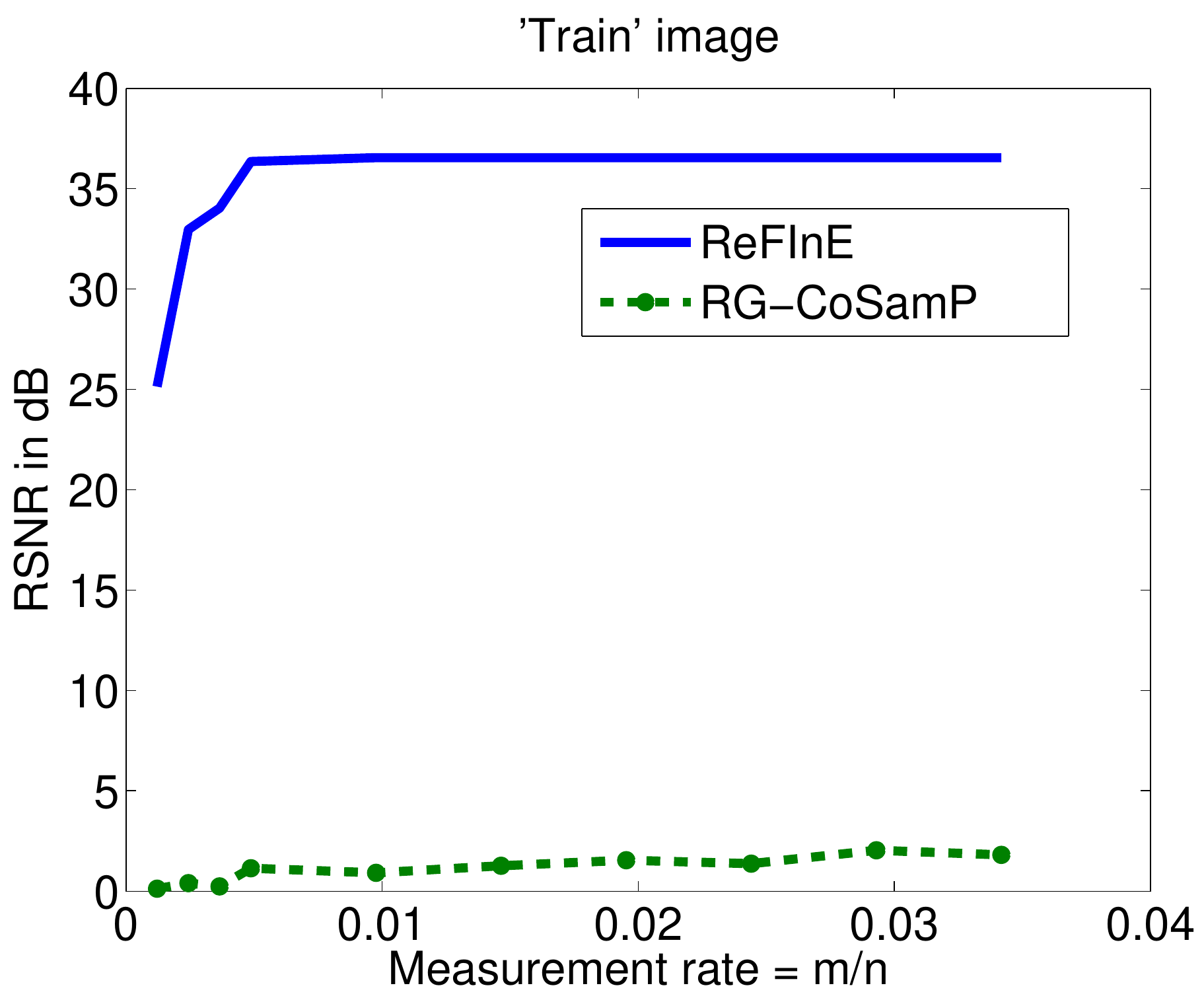}
				\label{fig:subfig13}
			}
			\subfigure[]{\includegraphics[height=14em,width=20em]{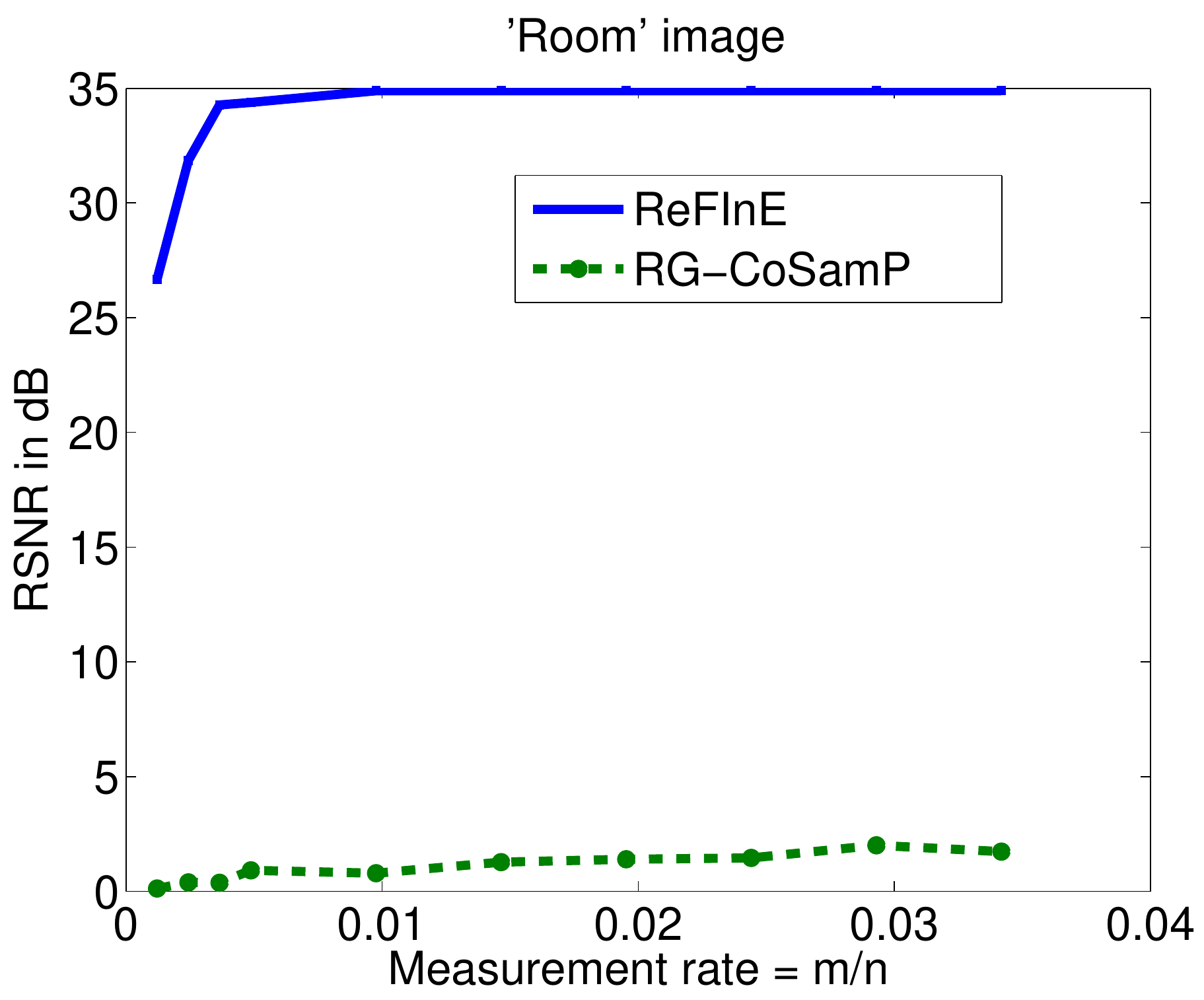}
				\label{fig:subfig14}
			}
			\caption{The figure shows variation of image-wise RSNR for recovered integral image estimates for the four images. It is very clear that for all the four images, our framework outperforms `RG-CoSamP' in terms of RSNR, at all measurement rates.}
			\label{fig:Integral}
		\end{figure*}

{{\bf Estimation of integral images:}}
		We show that good quality estimates of integral images can be obtained using our framework. To this end, we first construct {\bf ReFInE} measurement matrices of various ranks, $M$. We achieve this by considering the SVD of ${\bf Q}^*$ obtained above. For a particular value of $M$, the {\bf ReFInE} $\phi$ is calculated according to $\phi_{M} = {\bf \Sigma}_{M}^{1/2}{\bf V}_M^{T}$, where ${\bf \Sigma}_M = diag\{\lambda_1,..,\lambda_M\}$ is a diagonal matrix with $M$ largest singular values arranged along the diagonal and ${\bf V}_M^{T}$ denote the corresponding rows in ${\bf V}^{T}$. Its dual, $\phi^d$, is calculated similarly. For each particular measurement rate, determined by the value of $M$, the integral image estimates are recovered from $M$ {\bf ReFInE} measurements for all the 4952 test images in the PASCAL VOC 2007 dataset \cite{pascal-voc-2007}. Similarly integral image estimates are recovered from random Gaussian measurements by first performing non-linear iterative reconstruction using the CoSamP algorithm \cite{needell2009cosamp} and then applying the integral operation on the reconstructed images. This pipeline is used as baseline to compare integral estimates, and henceforth is referred to as `RG-CoSamP'. We then measure the recovered signal-to-noise ratio (RSNR) via 20 $\log_{10}\left(\frac{\norm{\hat{I}}_F}{\norm{\hat{I}-I}_F}\right)$.
		\begin{table*}
			\centering
			{\begin{tabular}{| l | l | l | l | l | l | l |}
					\hline
					Method & {\bf ReFInE} & RG-CoSamP & {\bf ReFInE} & RG-CoSamP &{\bf ReFInE} & RG-CoSamP \\ \hline
					$M$ (measurement ratio) & 20 (0.005) &  20 (0.005) & 40 (0.01) & 40 (0.01) & 60 (0.015) & 60 (0.015) \\ \hline
					Time in s & 0.0034 & 0.38 & 0.0036 & 0.58 & 0.0031  & 0.97  \\ \hline
					RSNR in dB & 38.95 & -16.76 & 38.96 & -11.22 & 38.96 & -10.9 \\ \hline 
				\end{tabular}
			}
			\caption{Comparison of average RSNR and time for recovered integral image estimates obtained using our method with RG-CoSamP. Our framework outperforms RG-CoSamP in terms of both recovery signal-to-noise ratio and time taken to estimate the integral image, at all measurement rates. }
			\label{tb:RSNR}
        \end{table*}
		The average RSNR for recovered integral image estimates as well as the time taken to obtain integral images are tabulated in the table \ref{tb:RSNR}. Our framework outperforms RG-CoSamP in terms of both recovery signal-to-noise ratio and time taken to estimate the integral image, at all measurement rates. This shows that {\bf ReFInE} $\phi$, the measurement matrices designed by our framework, facilitate faster and more accurate recovery of integral image estimates than the universal matrices. The average time taken to obtain integral image estimates in our framework is about 0.003s, which amounts to a real-time speed of ~300 FPS. Further, we randomly select four images (`Two Men', `Plane', `Train' and `Room') from the test set (shown in figure \ref{fig:twomen}, \ref{fig:plane}, \ref{fig:train}, \ref{fig:room}) and present qualitative and quantitative results for individual images.   
		Image-wise RSNR v/s measurement rate plots are shown in figure \ref{fig:Integral}. It is very clear that for all the four images, our framework clearly outperforms RG-CoSamP in terms of RSNR, at all measurement rates.
		
        \noindent
		\par
\begin{figure*}[ht!]
			\centering
			\subfigure[]{\includegraphics[height=14em,width=20em]{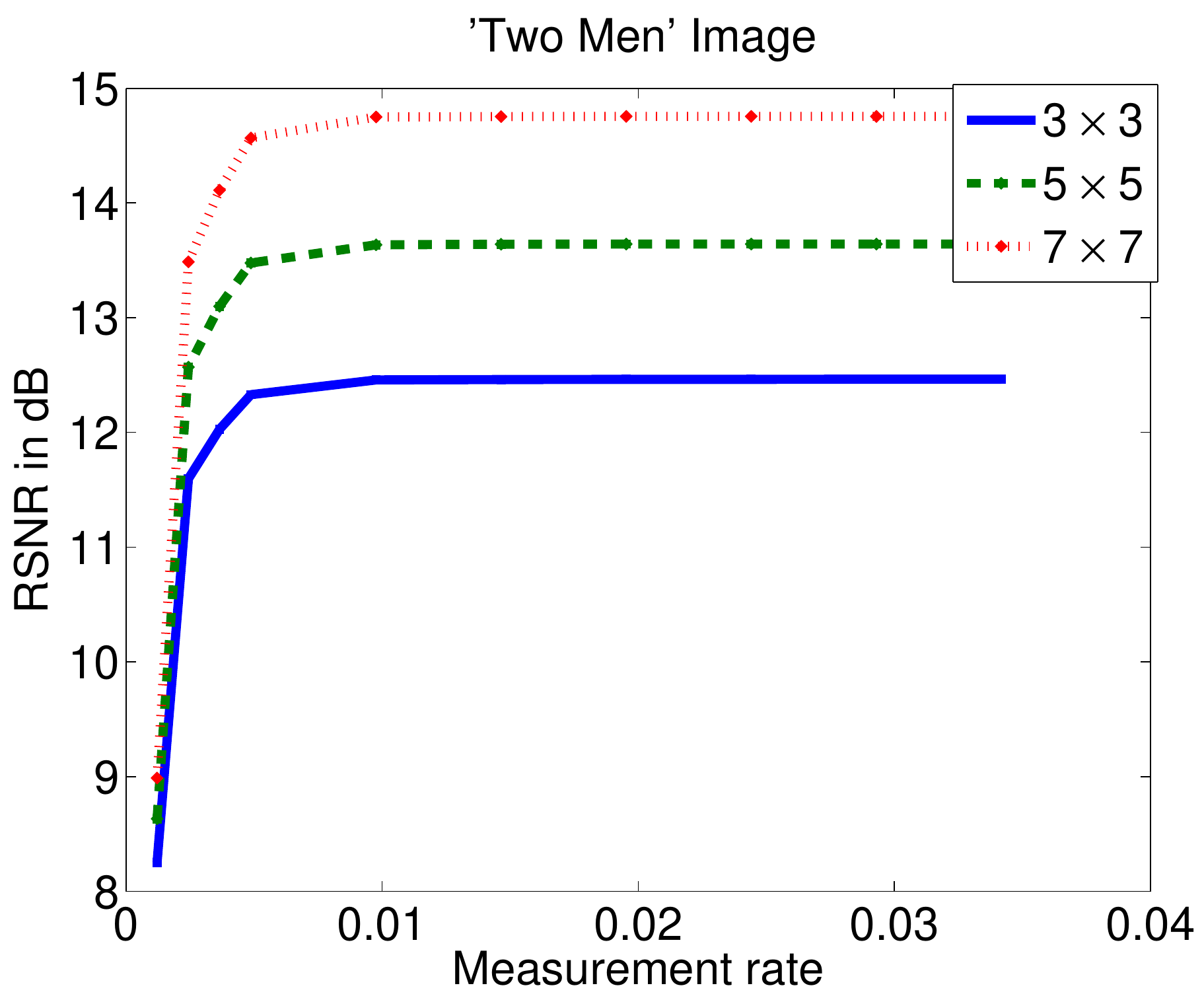}
				\label{fig:subfig1}
			}
			\subfigure[]{\includegraphics[height=14em,width=20em]{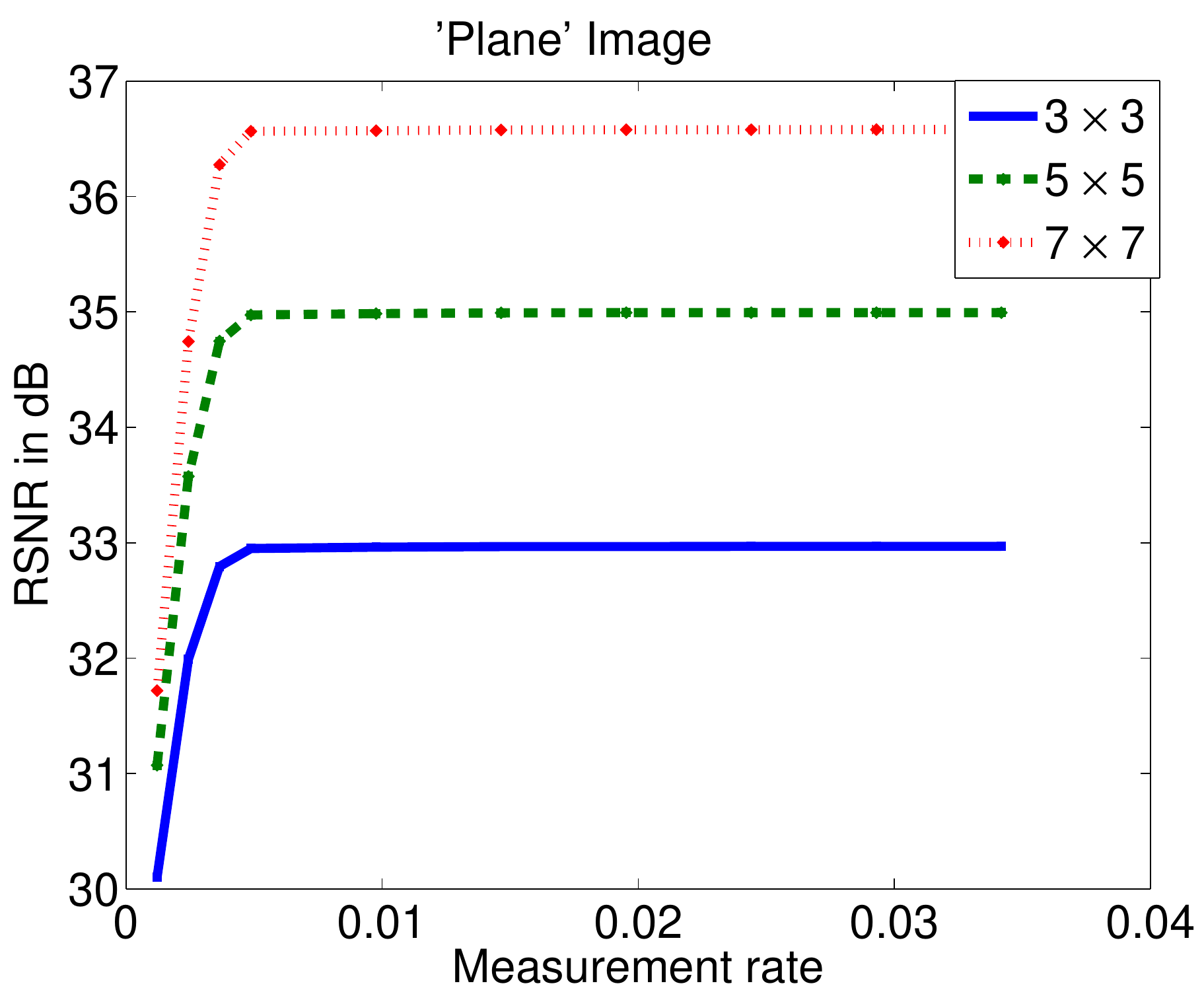}
				\label{fig:subfig2}
			}
			\subfigure[]{\includegraphics[height=14em,width=20em]{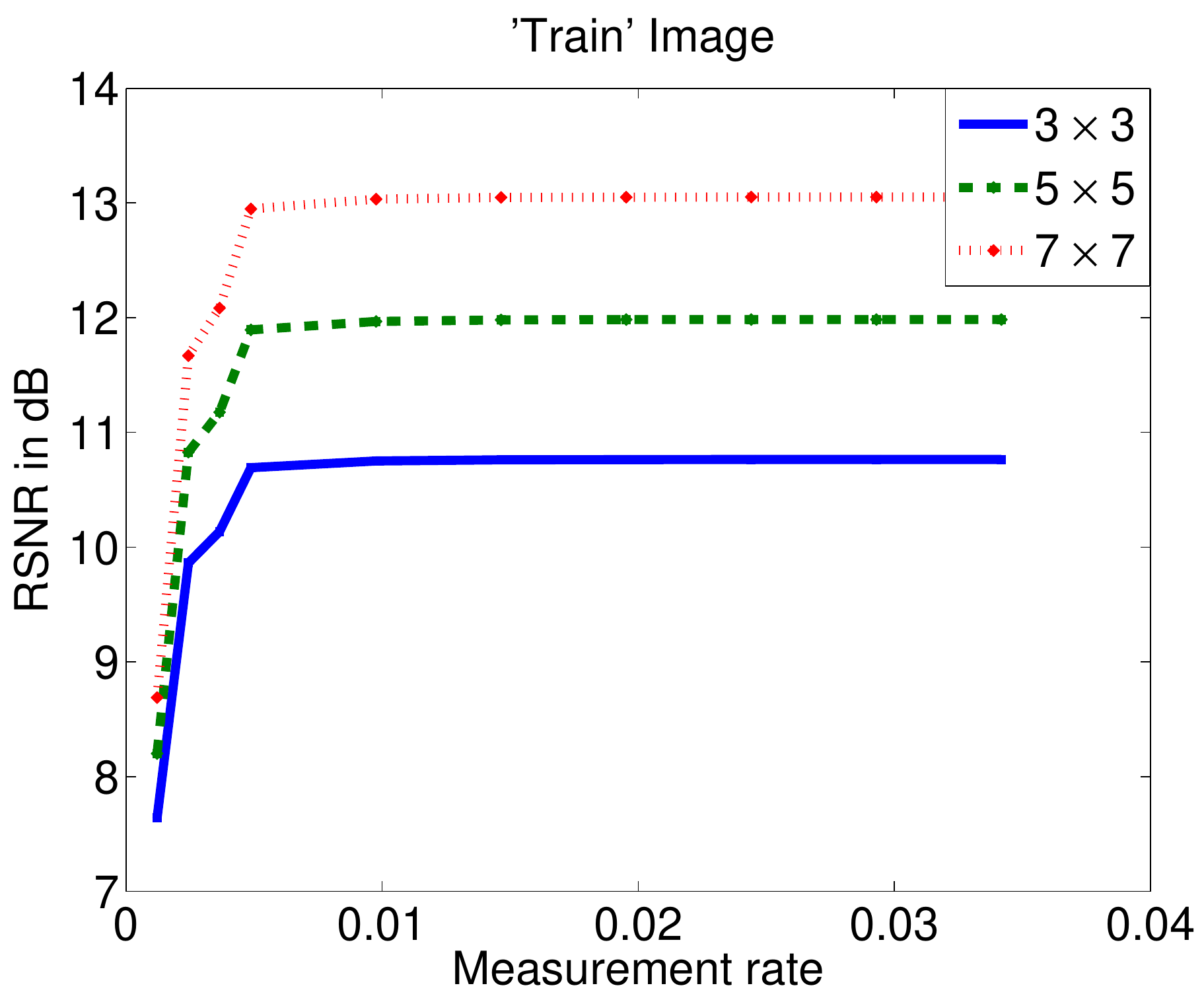}
				\label{fig:subfig3}
			}
			\subfigure[]{\includegraphics[height=14em,width=20em]{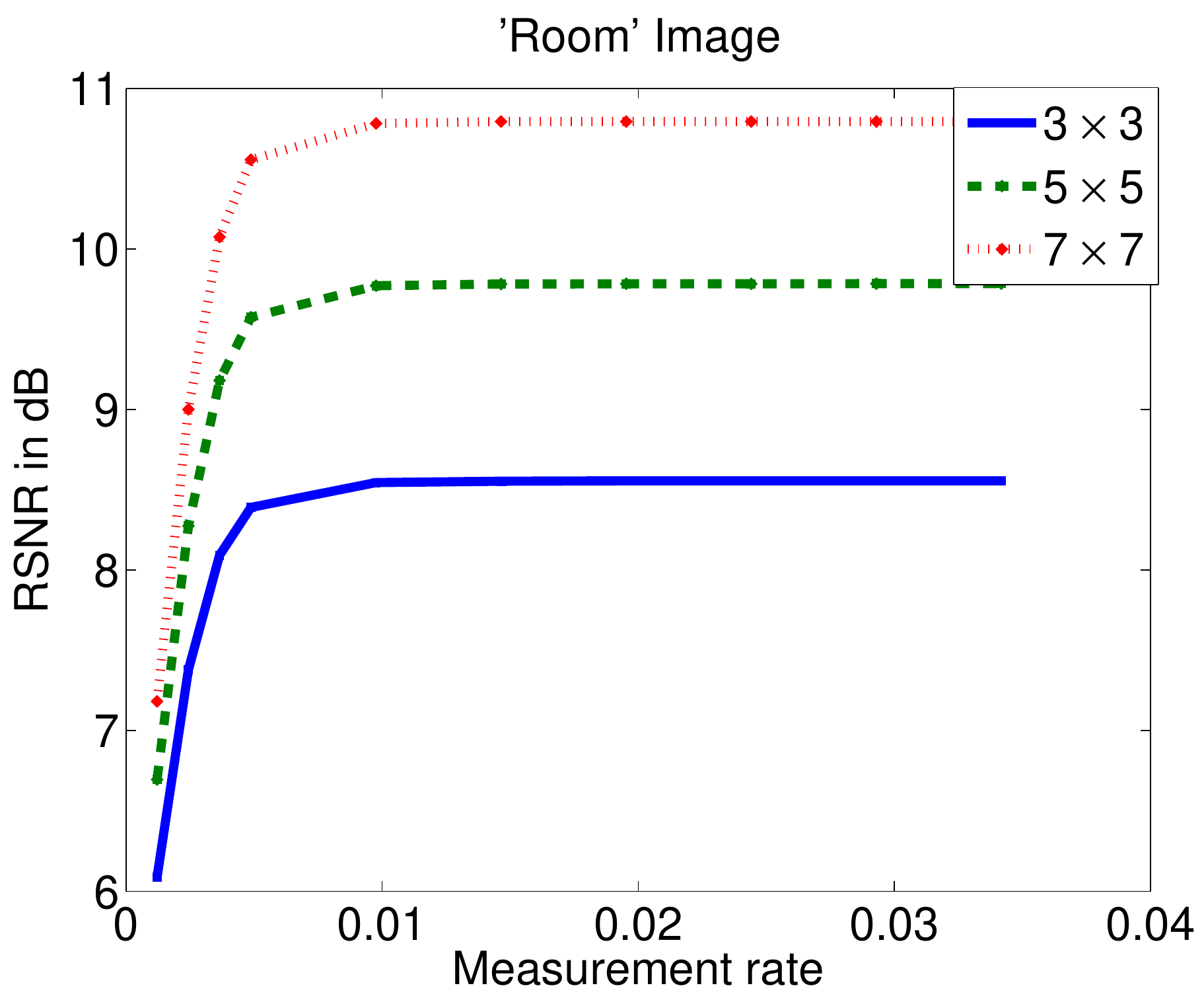}
				\label{fig:subfig4}
			}
			\caption{\small{The figure shows the variation of  RSNR for the recovered box-filtered outputs using {\bf ReFInE} with measurement rate. It is evident that even for 1\% measurement rate, we obtain high RSNR box-filtered outputs. For a fixed measurement rate, the RSNR increases with the size of the filter. This shows the structures global in nature are captured better. This is particularly true in the case of `Plane' image. The high RSNR for this image hints at the absence of fine structures and homogeneous background.}}
			\label{fig:PSNR1}
			
		\end{figure*}

{{\bf Estimation of box-filtered outputs:}}
		It is well known that box-filtered outputs of any size can efficiently computed using integral images \cite{viola2004robust}. To show the capability of our framework in recovering good quality box-filtered output estimates, we conduct the following experiment. For box filters of sizes $3 \times 3$, $5 \times 5$ and $7 \times 7$, we compute the estimates of filtered outputs for the four images using their respective recovered integral image estimates. RSNR v/s measurement rate plots for different filter sizes are shown in figure \ref{fig:PSNR1}. 
		It is evident that even for a remarkably low measurement rate of 1\% , we obtain high RSNR box-filtered outputs. For a fixed measurement rate, expectedly the RSNR increases with the size of the filter. This shows the structures which are more global in nature are captured better. This is particularly true in the case of `Plane' image. The high RSNR for this image hints at the absence of fine structures and homogeneous background. 
		\begin{figure*}[ht!]
			\centering
			\subfigure[$3 \times 3$]{\includegraphics[height=9em,width=40em]{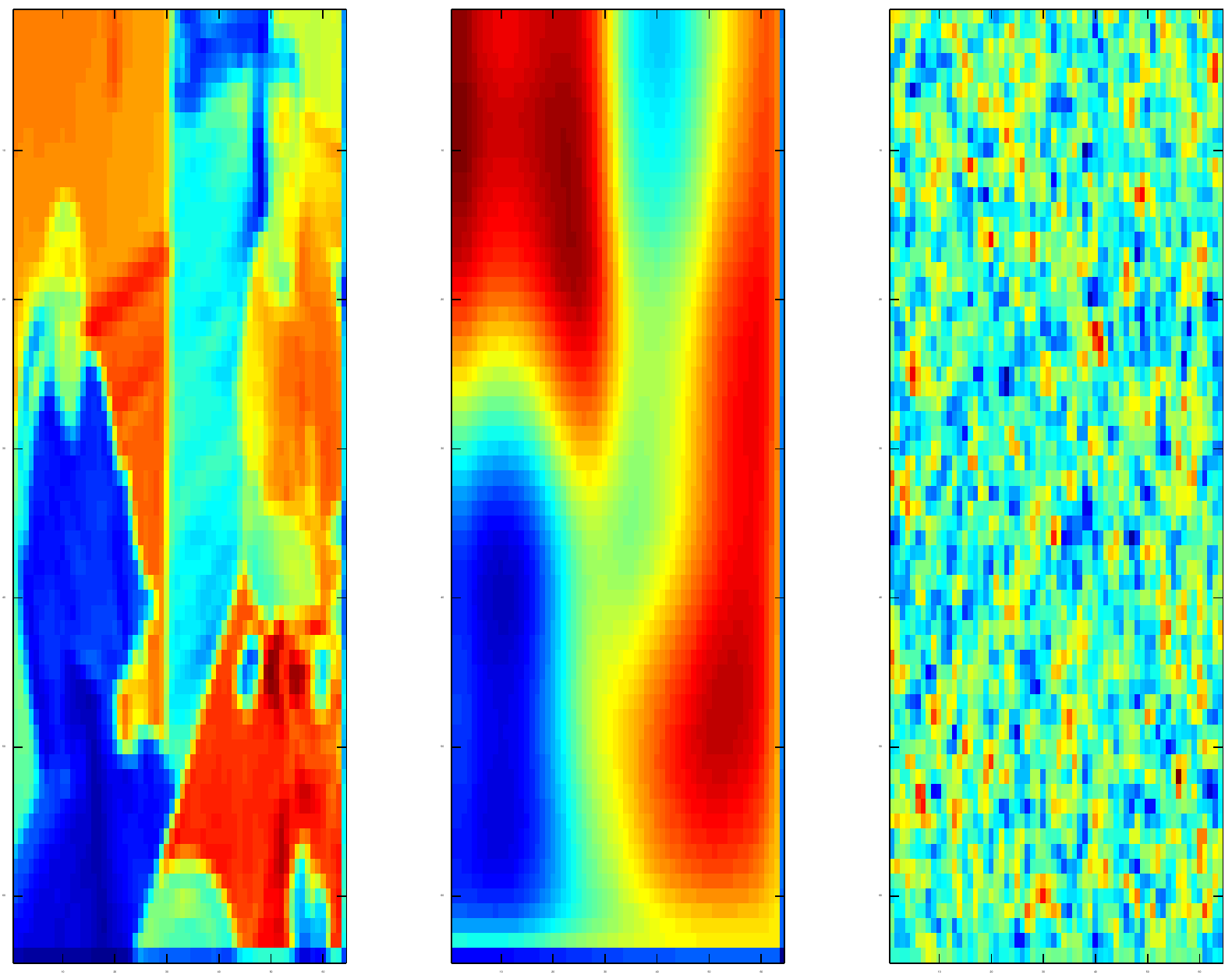}
				
				\label{fig:subfig5}
			} \\

			\subfigure[$7 \times 7$]{\includegraphics[height=9em,width=40em]{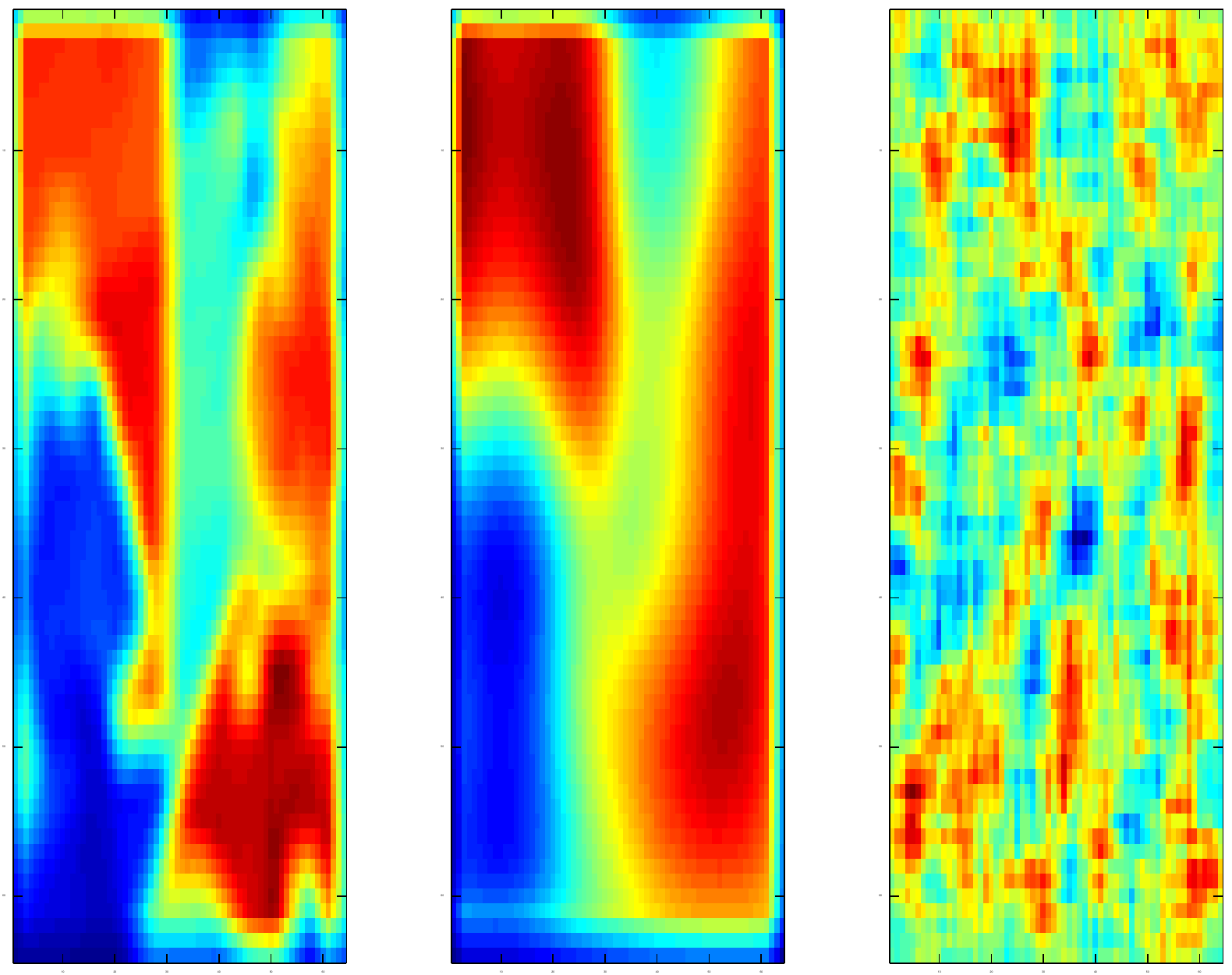}
				\label{fig:subfig6}
			}
			\caption{\small{Heat maps for box-filtered outputs for the `Two men' image. Left to right: Exact output, {\bf ReFInE} (m/n = 0.01), and RG-CosamP (m/n = 0.01). It is clear that greater quality box-filtered output estimates can be recovered from {\bf ReFInE} measurements and the  recovered outputs retain the information regarding medium-sized structures in the images, while in case of RG-CoSamP, the output is all noisy and does not give us any information.}}

			\label{fig:Box3}
		\end{figure*}
		Further, for the `Two Men' Image, we also compare the heat maps of the exact box-filtered outputs with the estimated ones. We fix the measurement rate to 1\%. For filter sizes $3 \times 3$ and $7 \times 7$, the exact box-filtered outputs are computed and compared with the box-filtered output estimates obtained using our framework, and RG-CoSamP as well. The heat map visualizations of the outputs are shown in figure \ref{fig:Box3}. It is clear that greater quality box-filtered output estimates can be recovered using our framework and the  recovered outputs retain the information regarding medium-sized structures in the images, while in the case of RG-CoSamP, the output is all noisy and does not give us any information.

		\section{Tracking Application}
		In this section, we show the utility of the framework in practical applications. In particular, we show tracking results on 50 challenging videos used in benchmark comparison of tracking algorithms \cite{wu2013online}. We emphasize that our aim is not to obtain state-of-the-art tracking results but to show that integral image estimates can be used to obtain robust tracking results at low measurement rates. To this end, we conduct two sets of tracking experiments, one with original resolution videos and one with high definition videos. 
		\par{{\bf Tracking with original resolution videos:}}
		We conduct tracking experiments on original resolution videos at three different measurement rates, viz 1.28\%, 1\%, and 0.49\%. In each case, we use the measurement matrix obtained for block size of $32 \times 32$, and obtain {\bf ReFInE} measurements for each frame using the $\phi^{*}$ obtained as above. Once, the measurements are obtained, our framework recovers integral image estimates from these measurements in real time. The estimates are subsequently fed into the best performing Haar-feature based tracking algorithm, Struck \cite{hare2011struck} to obtain the tracking results. Henceforth, we term this tracking pipeline as {\bf ReFInE+Struck}. To evaluate our tracking results, we use the standard criterion of precision curve as suggested by \cite{wu2013online}. To obtain the precision curve, we plot the precision scores against a range of thresholds. A frame contributes to the precision score for a particular threshold, $\alpha$ if the distance between the ground truth location of the target and estimated location by the tracker is less than the threshold, $\alpha$. Precision score for given threshold is calculated as the percentage of frames which contribute to the precision score. When precision scores are required to be compared with other trackers at one particular threshold, generally threshold is chosen to be equal to 20 pixels \cite{wu2013online}.
		\par
\begin{figure}[ht!]
		   	\centering
		   	\subfigure[]{\includegraphics[height=17em,width=20em]{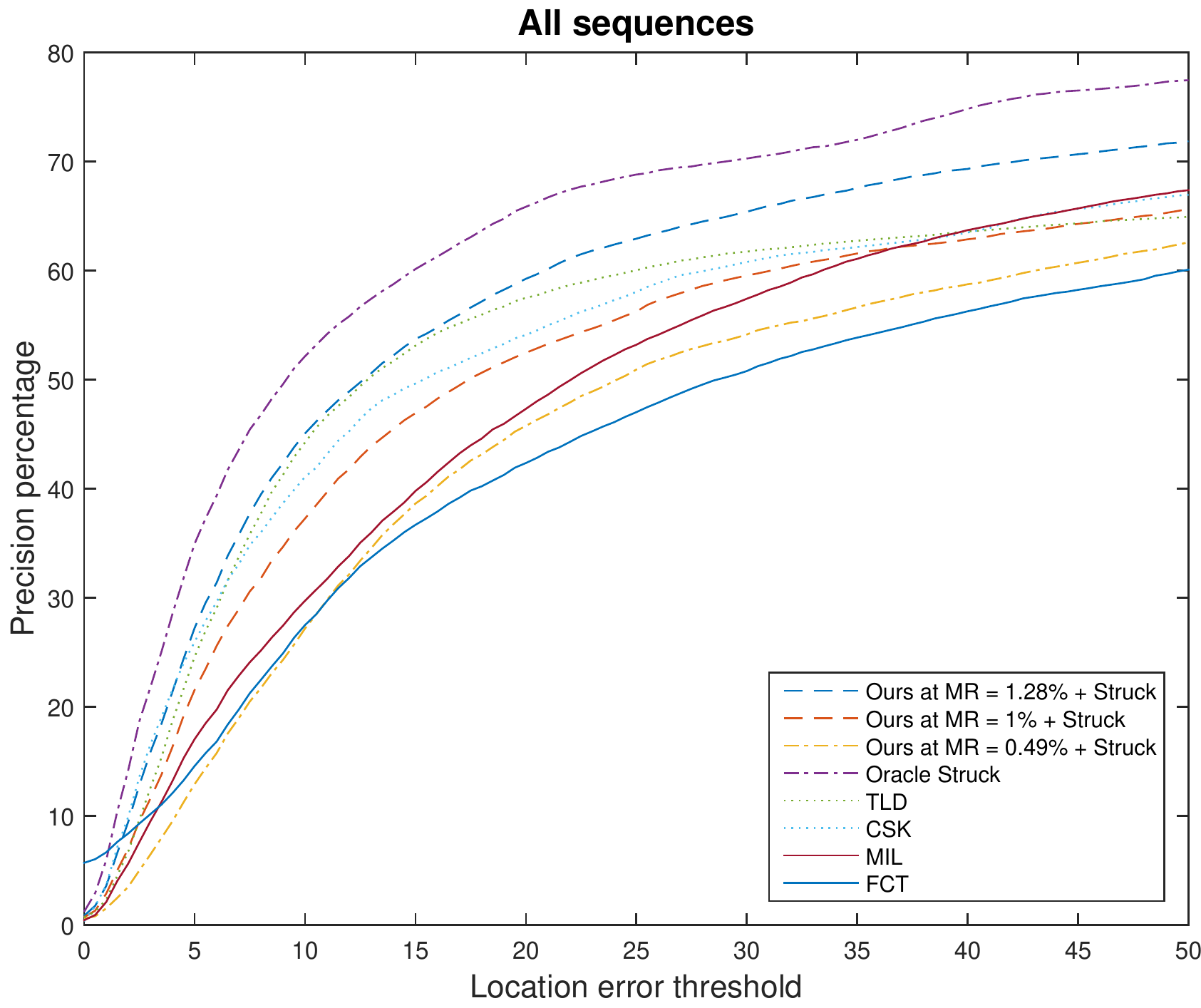}
		   		\label{fig:subfig1}
		   	}
		   	
		   	\caption{\small{ `ReFInE+Struck' at a measurement rate of 1.28\% performs better than other trackers, MIL, CSK, TLD, and FCT and only a few percentage points worse than Oracle Struck for all thresholds. Even at a measurement rate of 1\%, `ReFInE+Struck' performs nearly as well as TLD and CSK trackers which operate at 100\% measurement rate.}}
		   	\label{fig:pre_all}
		   \end{figure}

{{\bf Precision curve:}} 
		The precision curves for our framework at the three different measurement rates are plotted against a wide range of location error thresholds, and are compared with the same for other trackers, Oracle Struck \cite{hare2011struck}, and various other trackers, TLD \cite{kalal2010pn}, MIL \cite{babenko2011robust}, CSK \cite{henriques2012exploiting}, and FCT \cite{zhang2014fast} in figure \ref{fig:pre_all}.  It is to be noted all the trackers used for comparison utilize full-blown images for tracking and hence operate at 100\% measurement rate.  As can be seen clearly, `ReFInE+Struck' at 1.28\% performs better than other trackers, MIL, CSK, TLD, and FCT and only a few percentage points worse than Oracle Struck for all thresholds. In particular, the mean precision over all 50 sequences in the dataset \cite{wu2013online} for the threshold of 20 pixels is obtained for `ReFInE+Struck' at three different measurement rates and is compared with other trackers in table \ref{tb:Precision_or_resol}. We obtain a precision of 59.26\% at a measurement of 1.28\%, which is only a few percentage points less than precision of 65.5\% using Oracle Struck and 60.8\% using TLD. Even at an extremely low measurement rate of 0.49\%, we obtain mean precision of 45.78\% which is competitive when compared to other trackers, MIL, and FCT which operate at 100\% measurement rate. This clearly suggests that the small number of well-tailored measurements obtained using our framework retain enough information regarding integral images and hence also the Haar-like features which play a critical role in achieving tracking with high precision. 
		\par{{\bf Frame rate:}}
		Even though, our framework uses Struck tracker, the frame rates at which `ReFInE+Struck' operates are potentially less than the frame rate that can be obtained with Oracle Struck, and can even be different at different measurement rates. This is due to the fact that once the measurements are obtained for a particular frame, we first have to obtain an intermediate reconstructed frame before applying the integral operation. However, in the case of Oracle Struck, the integral operation is applied directly on the measured frame. The frame rate for `Our+Struck' at different measurement rates are compared with the frame rates for other trackers in table \ref{tb:Precision_or_resol}. However, as can be seen, the preprocessing operation to obtain the intermediate reconstructed frame barely affects the speed of tracking since the preprocessing step, being multiplication of small-sized matrices can be efficiently at nearly 1000 frames per second.

	     \begin{table}
				\begin{tabular}{|l|l|l|}
					\hline
					Tracker & Mean Precision & Mean FPS  \\
					\hline
					ReFInE at MR = 1.28\% + Struck & 59.26 & 19.61\\
					\hline
					ReFInE at MR = 1\% + Struck & 52.47 & 19.61\\
					\hline
					ReFInE at MR = 0.49\% + Struck & 45.78 & 19.62\\
					\hline 
					Oracle Struck \cite{hare2011struck} &  65.5 & 20\\
					\hline 
					TLD \cite{kalal2010pn} & 60.8 & 28 \\
					\hline 
					CSK \cite{henriques2012exploiting} &  54.11 & 362 \\
					\hline 
					MIL \cite{babenko2011robust}  & 47.5 & 38 \\
					\hline 
					FCT \cite{zhang2014fast} & 42.37 & 34.92\\
					\hline
				\end{tabular}
				\caption{\small{Mean precision percentage for 20 pixels error and mean frames per second for various state-of-the-art trackers are compared with our framework at different measurement rates. The precision percentages for our framework are stable even at extremely low measurement rates, and compare favorably with other trackers which operate at 100\% measurement rate, i.e utilize all the pixels in the frames.}}	
				\label{tb:Precision_or_resol}
			\end{table}

	\par
\begin{figure*}[ht!]
			\centering
			\subfigure[]{\includegraphics[height=17em,width=20em]{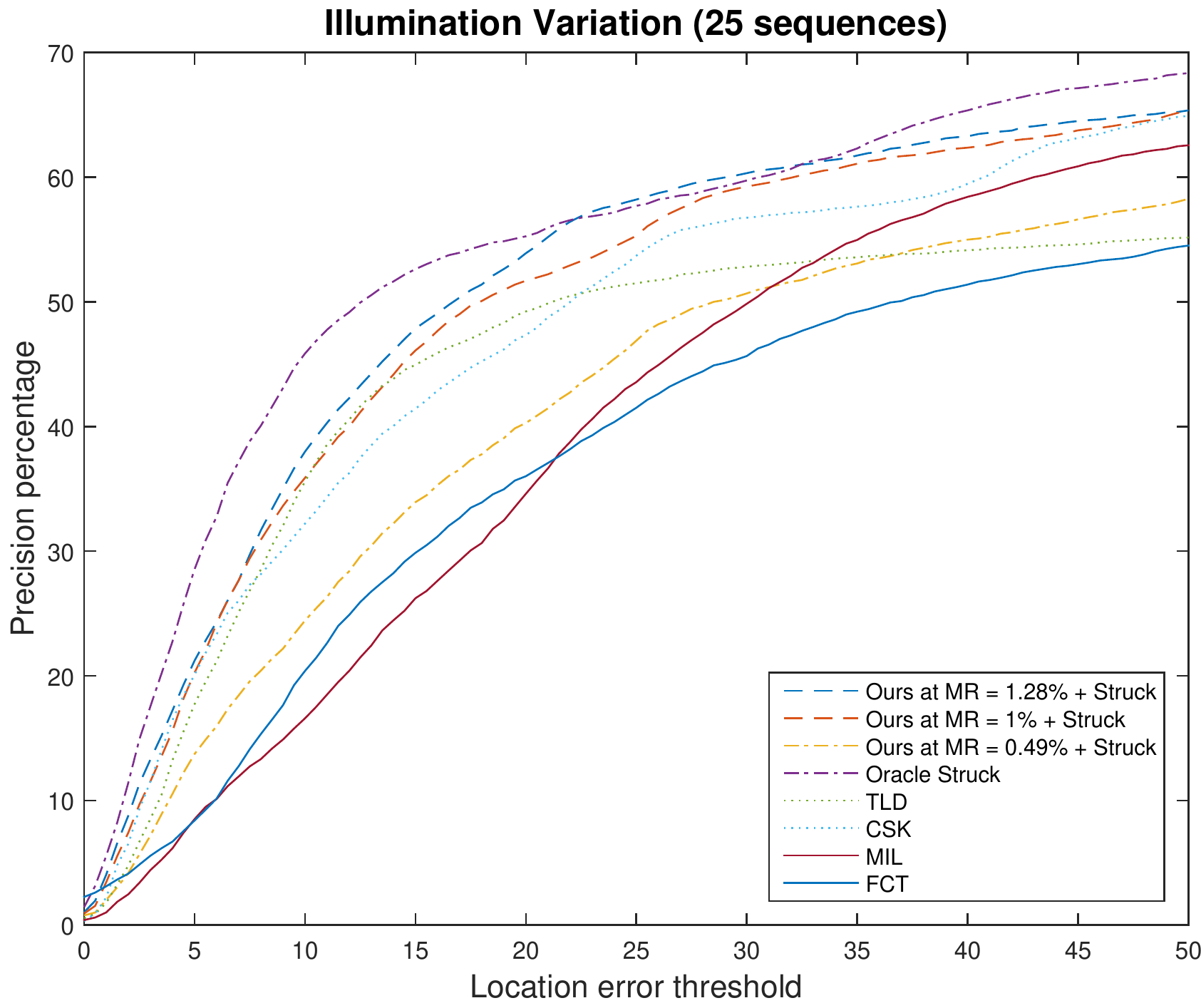}
				\label{fig:subfig1}
			}
			\hspace{0.4cm}
			\subfigure[]{\includegraphics[height=17em,width=20em]{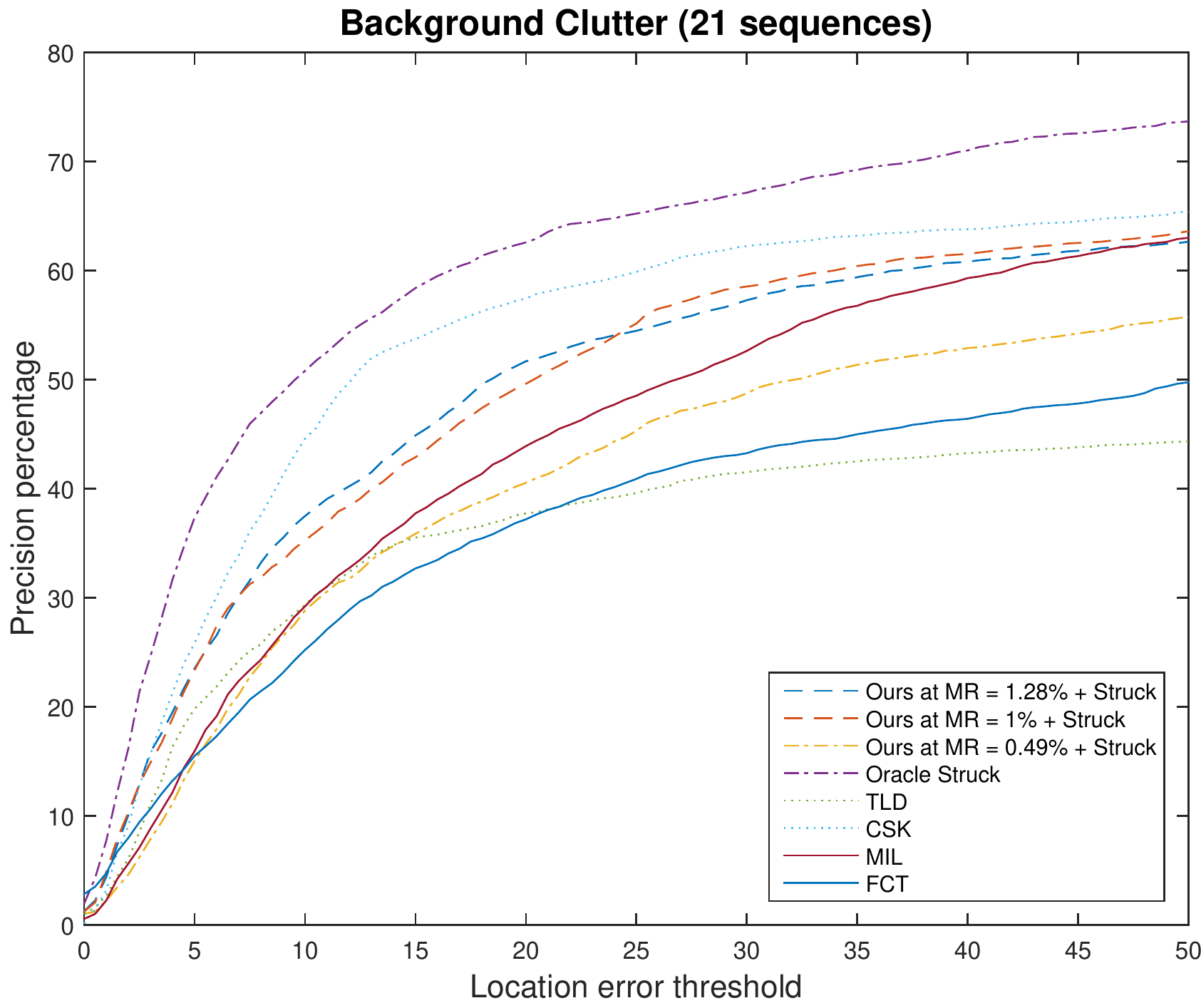}
				\label{fig:subfig2}
			}
			\\
			\subfigure[]{\includegraphics[height=17em,width=20em]{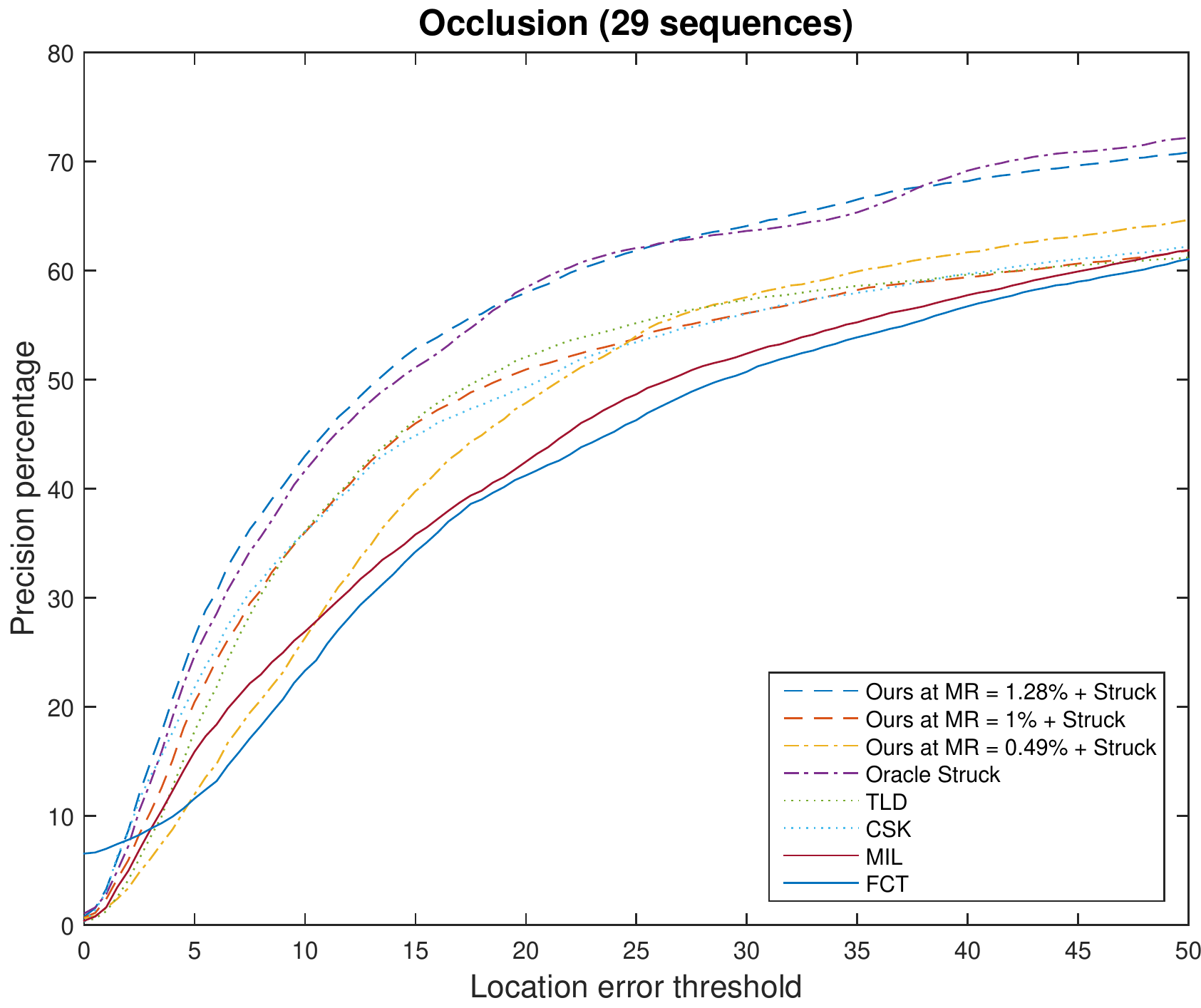}
				\label{fig:subfig3}
			}
			\hspace{0.4cm}
			\subfigure[]{\includegraphics[height=17em,width=20em]{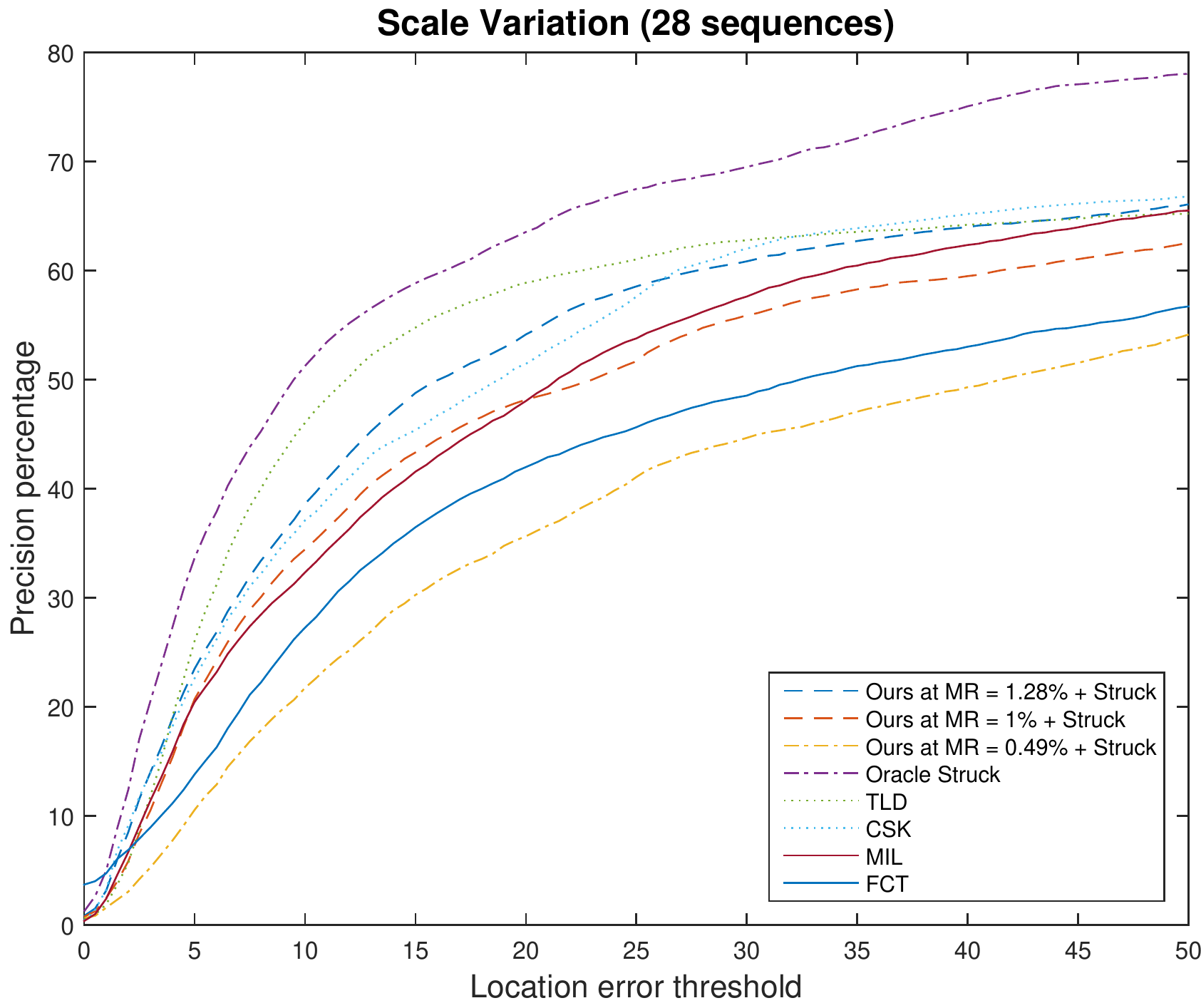}
				\label{fig:subfig4}
			}
			\caption{\small{Precision plots for four different attributes. In the case of `Illumination Variation' and `Occlusion' `ReFInE+Struck' at measurement rate of 1.28\% performs better than TLD, CSK, MIL and FCT, whereas in the case of the `Background Clutter' and `Scale Variation' attributes, TLD performs slightly better than `ReFInE+Struck' at measurement rate of 1.28\%.}}
			\label{fig:pre_ill_bac_occ_sca}
			
		\end{figure*}

{{\bf Experiments with sequence attributes:}}
	Each video sequence in the benchmark dataset is annotated with a set of attributes, indicating the various challenges the video sequence offers in tracking. We plot precision percentages against the location error threshold for each of these 10 different kinds of attributes. Figure \ref{fig:pre_ill_bac_occ_sca} shows the corresponding plots for attributes, `Illumination Variation', `Background Clutter', `Occlusion', and `Scale Variation'.  In the case of `Illumination Variation' and `Occlusion' `Our+Struck' at measurement rate of 1.28\% performs better than TLD, CSK, MIL and FCT, whereas in the case of the `Background Clutter' and `Scale Variation' attributes, TLD performs slightly better than `Our+Struck' at measurement rate of 1.28\%.

\begin{figure*}[ht!]
	\centering
	\subfigure[]{\includegraphics[height=17em,width=20em]{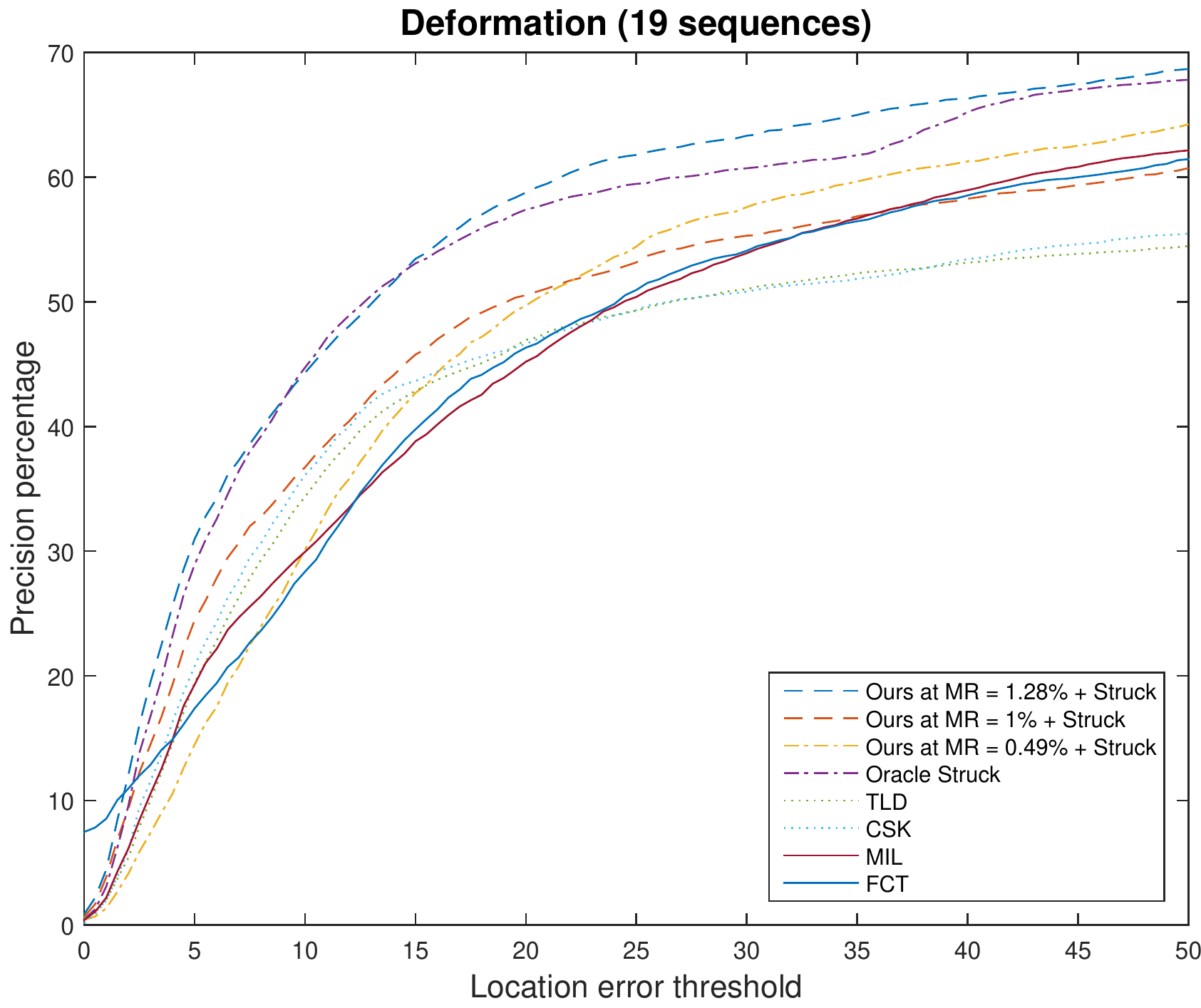}
		\label{fig:subfig1}
	}
	\hspace{0.4cm}
	\subfigure[]{\includegraphics[height=17em,width=20em]{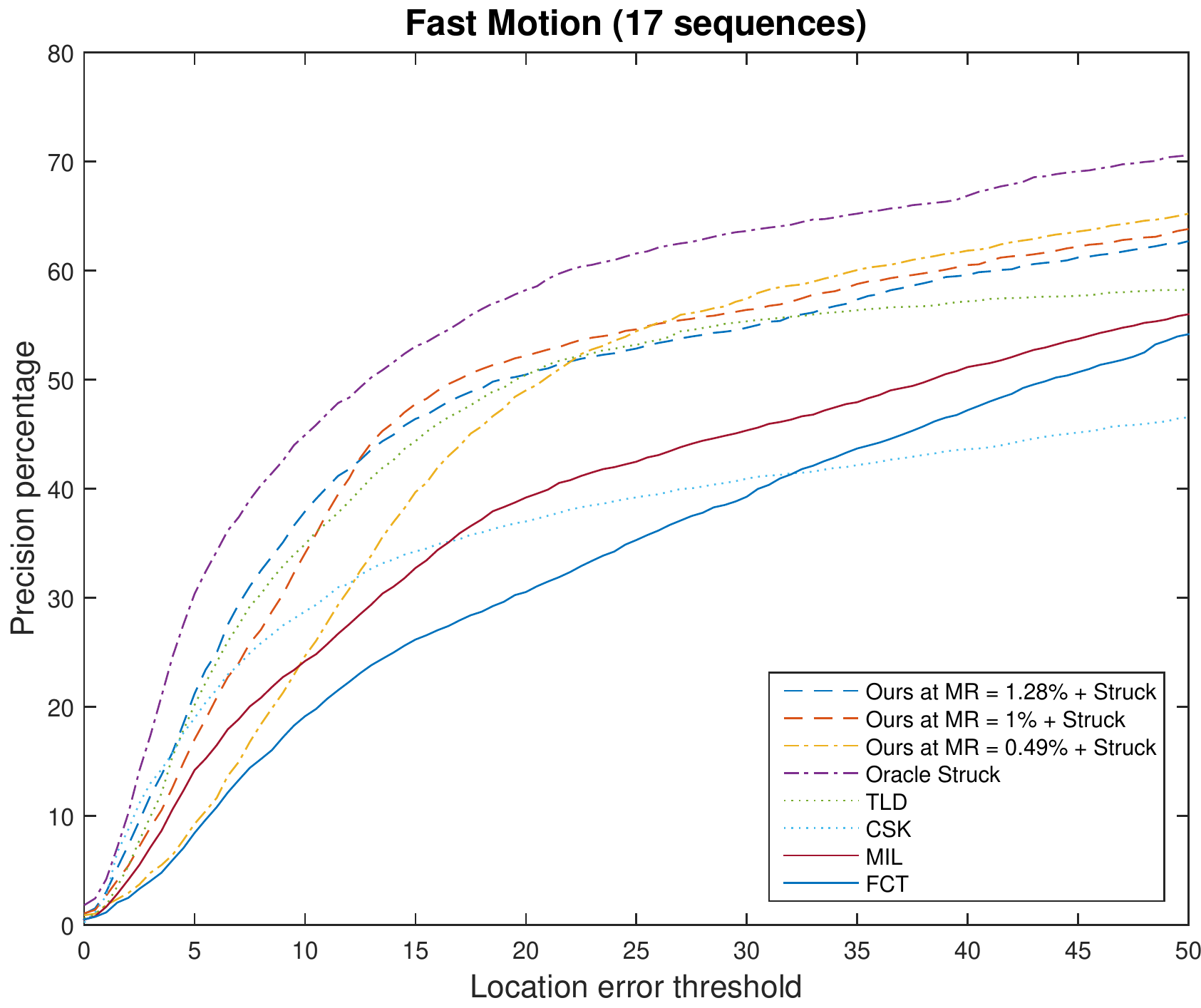}
		\label{fig:subfig2}
	}
	\\
	\subfigure[]{\includegraphics[height=17em,width=20em]{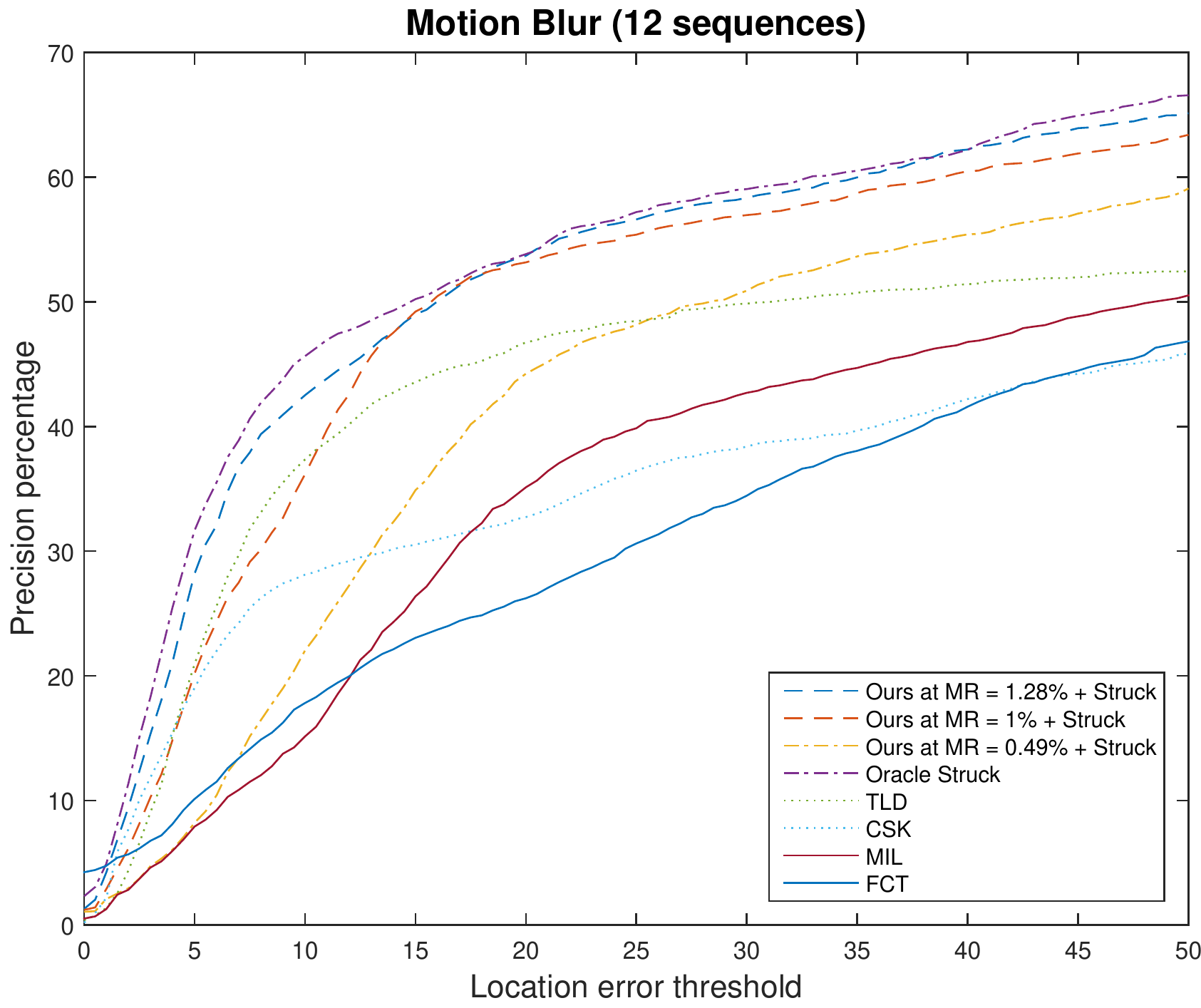}
		\label{fig:subfig3}
	}
	\hspace{0.4cm}
	\subfigure[]{\includegraphics[height=17em,width=20em]{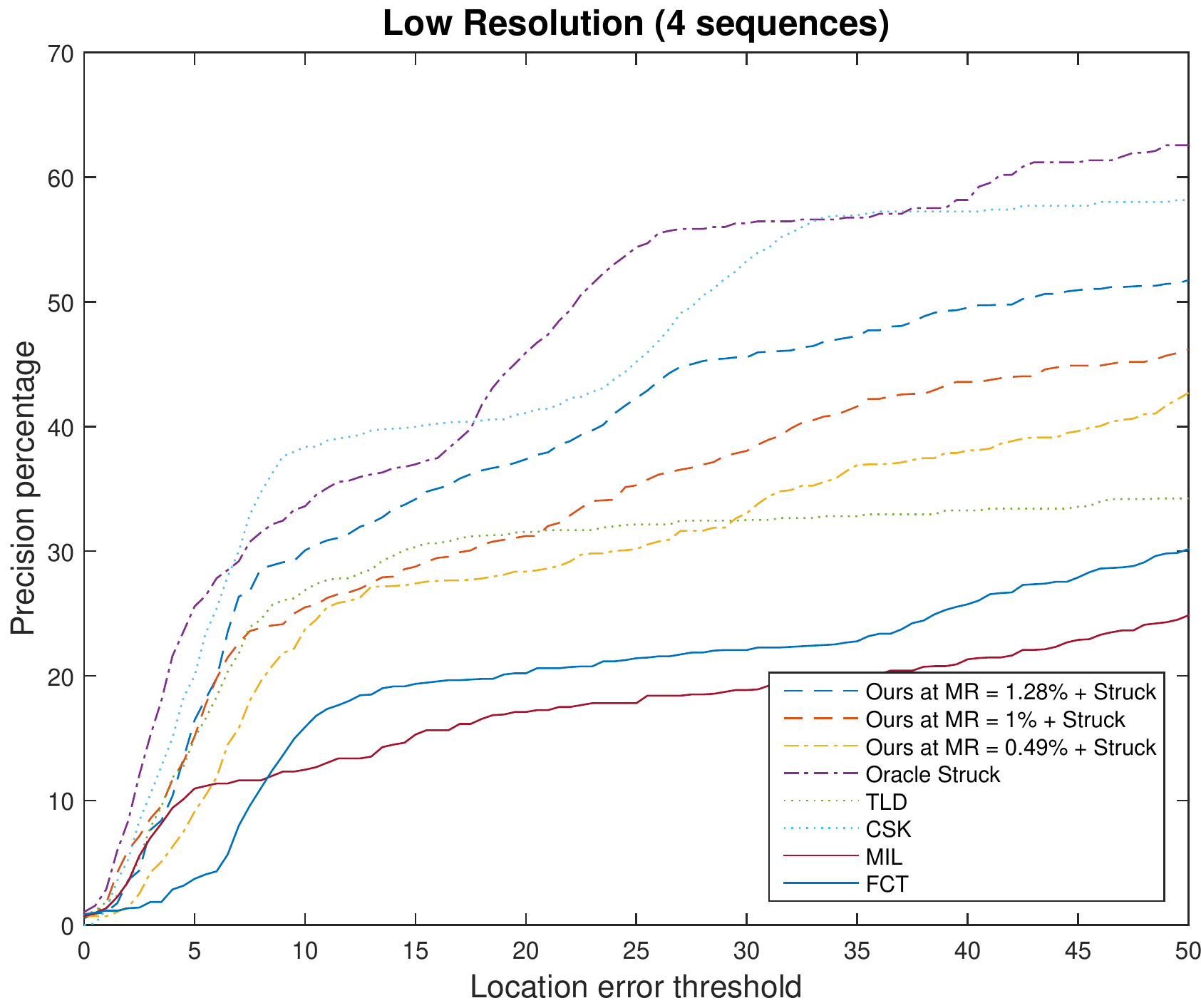}
		\label{fig:subfig4}
	}
	\caption{\small{Precision plots for four different attributes. In the cases of  `Deformation', `Fast Motion' and `Motion Blur', `ReFInE+Struck' at measurement rate of 1.28\% performs better than TLD, CSK, MIL and FCT, whereas in the case of `Low Resolution', TLD performs better than `ReFInE+Struck'.}}
	\label{fig:pre_def_fm_mb_lr}
\end{figure*}

Figure \ref{fig:pre_def_fm_mb_lr} shows the corresponding plots for attributes, `Deformation', `Fast Motion', `Motion Blur', and `Low Resolution'. In the cases of  `Deformation', `Fast Motion' and `Motion Blur', `ReFInE+Struck' at measurement rate of 1.28\% performs better than TLD, CSK, MIL and FCT, whereas in the case of `Low Resolution', TLD performs better than `ReFInE+Struck'.

\begin{figure*}[ht!]
			\centering
			\subfigure[]{\includegraphics[height=14em,width=15em]{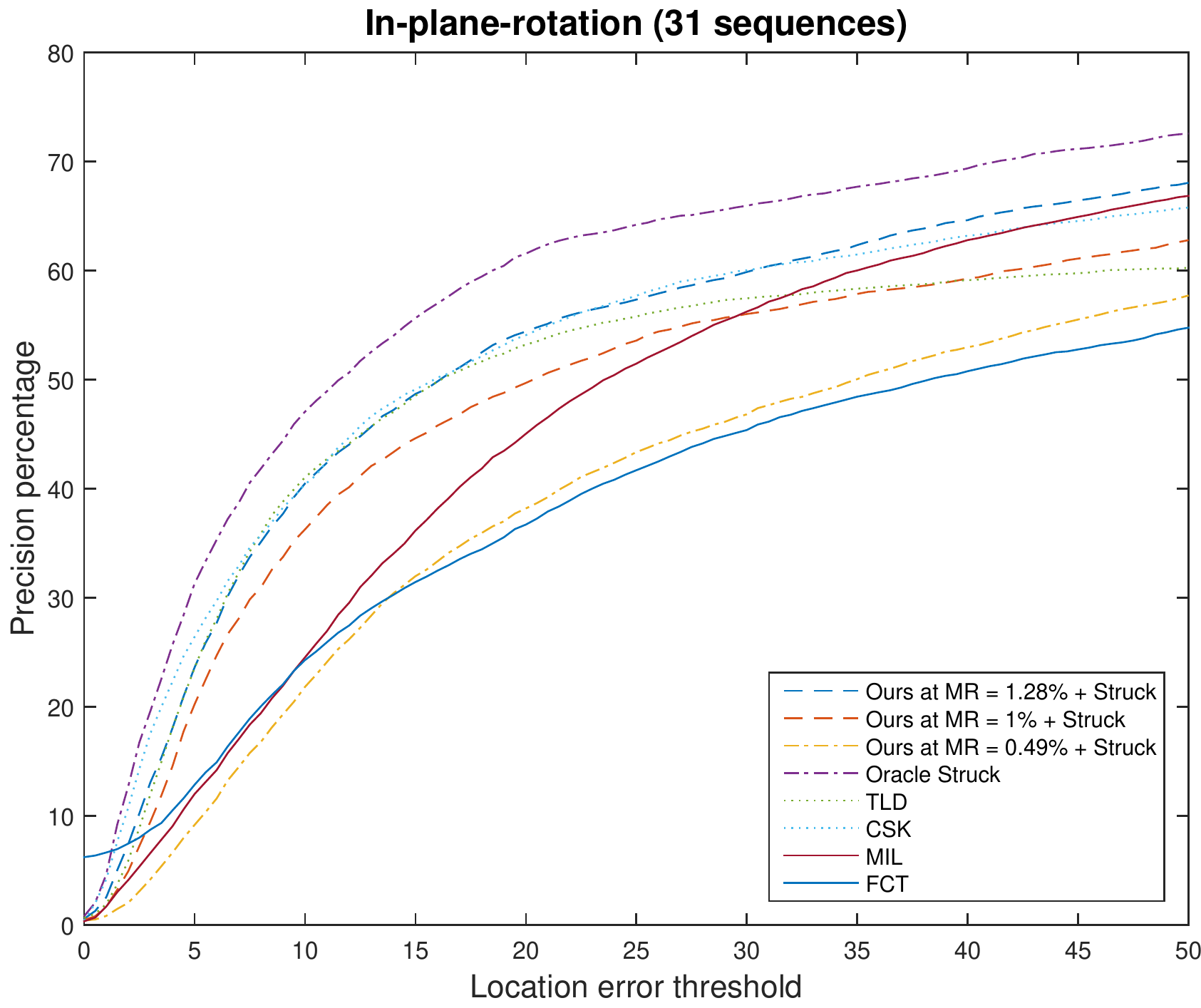}
				\label{fig:subfig1}
			}
			\hspace{0.2cm}
			\subfigure[]{\includegraphics[height=14em,width=15em]{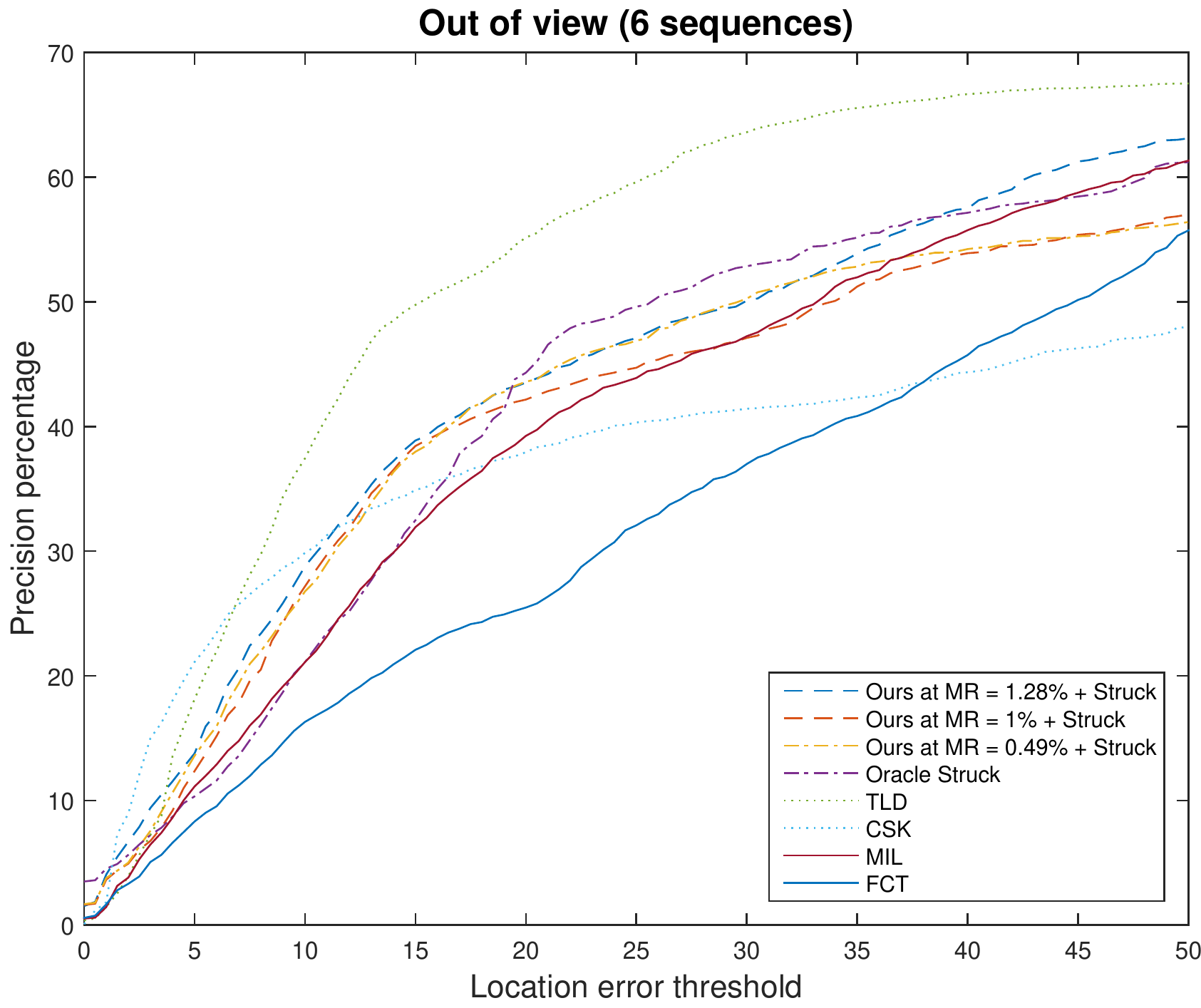}
				\label{fig:subfig2}
			}
	     	\hspace{0.2cm}
				\subfigure[]{\includegraphics[height=14em,width=15em]{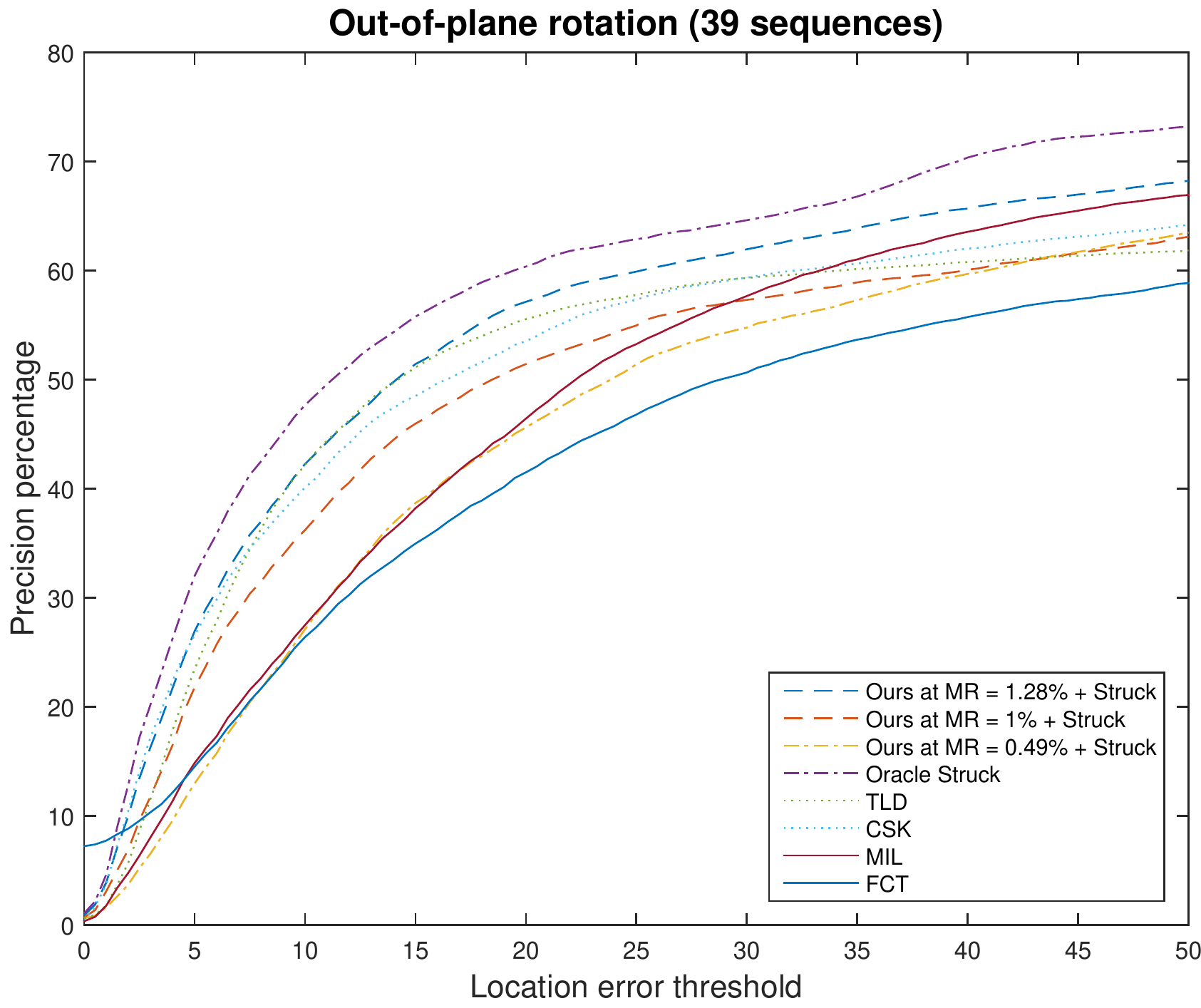}
					\label{fig:subfig2}
				}
			\caption{\small{Precision plots for three different attributes. In the cases of  `In the plane rotation', and `Out of plane rotation', `ReFInE+Struck' at measurement rate of 1.28\% performs better than TLD, CSK, MIL and FCT, whereas in the case of `Out of View', TLD performs better than `ReFInE+Struck'. }}
			\label{fig:pre_in_out_opr}
	\end{figure*}

Figure \ref{fig:pre_in_out_opr} shows the corresponding plots for attributes, `In the Plane rotation', `Out of View', and `Out of Plane rotation'.  

	\par{{\bf Tracking with high resolution videos:}} Tracking using high-resolution videos can potentially lead to improvement in performance due to availability of fine-grained information regarding the scene. However, in many applications, the deployment of  high-resolution sensors is severely limited by the lack of storage capacity. In such scenarios, it will be interesting to see if the small number of ReFInE measurements of high-resolution videos can yield better tracking performance than the full-blown low-resolution videos.    To conduct tracking experiments on high resolution videos, we first employ a deep convolutional network based image super resolution (SR) algorithm, SRCNN, \cite{dong2014learning} to obtain high resolution frames of the 50 videos considered earlier in the section. The aspect ratio for all frames is maintained, and the upscaling factor for image super resolution is calculated such that the resolution of the longer dimension in the higher resolution frame is at least 1000 pixels. We found that upscale factors varies between 2 and 8 for various videos in the dataset. Once the high resolution videos are obtained, we proceed to obtain ReFInE measurements as before. We conduct tracking experiments at four different measurement rate (1\%, 0.49\%, 0.29\%, 0.2\%). Note that these different measurement rates are with respect to (wrt) the high-resolution frames, and the measurement rate wrt original resolution, which we call effective measurement rate (EMR), is given by the ratio of the number of ReFInE measurements per frame to the number of pixels in a frame of original resolution video. Here, the tracking algorithm, Struck which we used for original resolution videos does not scale well in terms of computational complexity. For higher resolution videos, where the search space is much larger, we found that Struck is too slow for real-time application. Instead, we use a faster Haar feature based tracking algorithm, FCT \cite{zhang2014fast} algorithm. Henceforth, we dub this tracking pipeline as {\bf SR+ReFInE+FCT}. Once tracking results are obtained for the high resolution videos are obtained, we normalize the coordinates so as to obtain the tracking outputs with respect to original resolution videos. The precision score is calculated as before. The mean precision percentage for 20 pixels error and mean frames per second for `SR + ReFIne + FCT' at various measurement rates are given in table \ref{tb:Precision_hi_resol} and are compared for the same for `Oracle FCT' which operates for full-blown original resolution videos. It is clear that we obtain a significant boost in tracking accuracy for high resolution videos. At measurement rate of 1\% (EMR of 8.16\%), we obtain a mean precision percentage of 54.83, which is 12.46 percentage points more than that for `Oracle FCT'. Even at a measurement rate of 0.2\% ((EMR of 1.63\%)), the precision percentage of 45.79, which is about 3.42 percentage points more than that for `Oracle FCT'. However, the more accurate precision comes at the cost of frame rate. Since the search space is much larger for high resolution videos, the speed of tracking for high resolution videos, is only about 20 FPS, while `Oracle FCT' operates at 34.92 FPS. But 20 FPS suffices for near real-time implementations.
    
   \begin{table}
   	\begin{tabular}{|l|l|l|}
   		\hline
   		\scriptsize{Tracker} & \scriptsize{Mean Precision} & \scriptsize{Mean FPS}  \\
   		\hline
   	\scriptsize{SR + ReFInE at EMR = 8.16\% + FCT} & 54.83 & 19.61\\
   		\hline
   	\scriptsize{SR + ReFInE at EMR = 4\% + FCT} & 53.03 & 19.61\\
   		\hline
   	\scriptsize{SR + ReFInE at EMR = 2.37\% + FCT} & 50.9 & 19.62\\
   		\hline 
\scriptsize{SR + ReFInE at EMR = 1.63\% + FCT} & 45.79 & 19.62\\
        \hline
\scriptsize{Oracle FCT \cite{zhang2014fast}} & 42.37 & 34.92\\
       \hline
  	\end{tabular}
   	\caption{\small{Mean precision percentage for 20 pixels error and mean frames per second for `SR + ReFIne + FCT' at various measurement rates are compared with `Oracle FCT'. Even at extremely low measurement rates, the precision percentages for `SR + ReFIne + FCT' are better that for `Oracle FCT' which operates on full-blown original resolution images.}}	
   	\label{tb:Precision_hi_resol}
   \end{table}  

		\section{Conclusions}
		In this paper, we qualitatively and quantitatively showed that it is possible obtain high quality estimates of integral images and box-filtered outputs directly from a small number of specially designed spatially multiplexed measurements called {\bf ReFInE} measurements. To show the practical applicability of the integral image estimates, we presented impressive reconstruction-free tracking results on challenging videos at an extremely low measurement rate of 1\%. We also showed that with only a small number of {\bf ReFInE} measurements on high-resolution videos, which is only a fraction (2-8\%) of the size of the original resolution, one can obtain significantly better object tracking results than using full blown original resolution videos. From a philosophical point of view, this points to the possibility of attaining greater performance on other computer vision inference tasks from a small number of carefully tailored spatially multiplexed measurements of high-resolution imagery rather than full-blown low resolution imagery.

\begin{appendices}
\section{Derivation: rank({\bf P}) = rank({\bf Q}) - 1}\label{app:rank}
By construction we have $(\phi^d)^T = [{\bf 1} | (\phi^d)_{2:m}^T] = [{\bf 1} | ({\bf D} {\bf U}_{2:n}^{T})^T]$. The column of all ones in $(\phi^d)^T$ is orthogonal to remaining $m-1$ columns. Hence, we have $rank(\phi_{2:m}^d) = rank(\phi^d) - 1 = rank({\bf Q}) - 1$. Similarly, we have $rank(\phi_{2:m}) = rank(\phi) - 1 = rank({\bf Q}) - 1$. Since, ${\bf P}$ is the product of equally ranked matrices, $(\phi^d)_{2:m}^T$ and $\phi$, it follows that  $rank({\bf P}) = rank(\phi^d_{2:m}) = rank({\bf Q}) - 1$.  
\section{Calculation of $\mathcal{A}_i^*({\bf y_i}^{k-1})$}\label{app:adjoint}
For brevity, we drop the superscript, $k - 1$. Let ${\bf \Sigma}^{1/2}_{w_d}{\bf U}_{2:n}^T = {\bf\Sigma}_{U}$. Consider the following equation.
\begin{equation}
 \langle \mathcal{A}({\bf P}), {\bf y} \rangle = \sum_{i=1}^{n}\langle {\bf P} , \mathcal{A}_i^*({ {\bf y_i}}) \rangle.
\label{eq:calcA1}
\end{equation} The left hand side can be written as 
\begin{align}
[{\bf y_1},...,{\bf y_n}][(\mathcal{A}_1({\bf P}))^T,...,(\mathcal{A}_n({\bf P}))^T]^T \\
= [{\bf y_1},...,{\bf y_n}][h_1^T {\bf P}^T {\bf\Sigma}_U^T,...,h_n^T {\bf P}^T {\bf\Sigma}_U^T]^T \\
= \sum_{i=1}^n {\bf y_i} {\bf\Sigma}_U {\bf P}^T h_i \\
= \sum_{i=1}^n \langle {\bf P}, {\bf\Sigma}_U^T {\bf y_i} h_i^T \rangle. 
\label{eq:calcA2}
\end{align}
Comparing the equations \ref{eq:calcA1} and \ref{eq:calcA2}, we have $\mathcal{A}_i^*({\bf y_i}) = {\bf\Sigma}_U^T {\bf y_i} h_i^T$.    

\end{appendices}
		
		
		%


		\ifCLASSOPTIONcompsoc
		\section*{Acknowledgments}
		\else
		\section*{Acknowledgment}
		\fi

		\ifCLASSOPTIONcaptionsoff
		\newpage
		\fi
		
		
		
		
		\bibliographystyle{IEEEtran}
		%
		\bibliography{PAMI_STSF_3}

\begin{thebibliography}{10}
\providecommand{\url}[1]{#1}
\csname url@samestyle\endcsname
\providecommand{\newblock}{\relax}
\providecommand{\bibinfo}[2]{#2}
\providecommand{\BIBentrySTDinterwordspacing}{\spaceskip=0pt\relax}
\providecommand{\BIBentryALTinterwordstretchfactor}{4}
\providecommand{\BIBentryALTinterwordspacing}{\spaceskip=\fontdimen2\font plus
\BIBentryALTinterwordstretchfactor\fontdimen3\font minus
  \fontdimen4\font\relax}
\providecommand{\BIBforeignlanguage}[2]{{%
\expandafter\ifx\csname l@#1\endcsname\relax
\typeout{** WARNING: IEEEtran.bst: No hyphenation pattern has been}%
\typeout{** loaded for the language `#1'. Using the pattern for}%
\typeout{** the default language instead.}%
\else
\language=\csname l@#1\endcsname
\fi
#2}}
\providecommand{\BIBdecl}{\relax}
\BIBdecl

\bibitem{nayar2006programmable}
S.~K. Nayar, V.~Branzoi, and T.~E. Boult, ``Programmable imaging: Towards a
  flexible camera,'' \emph{Intl. J. Comp. Vision}, vol.~70, no.~1, pp. 7--22,
  2006.

\bibitem{SPC}
{M.B. Wakin, J.N. Laska, M.F. Duarte, D. Baron, S. Sarvotham, D. Takhar, K.F.
  Kelly and R.G. Baraniuk}, ``An architecture for compressive imaging,'' in
  \emph{{IEEE} Conf. Image Process.}, 2006.

\bibitem{sankaranarayanan2012cs}
A.~C. Sankaranarayanan, C.~Studer, and R.~G. Baraniuk, ``Cs-muvi: Video
  compressive sensing for spatial-multiplexing cameras,'' in
  \emph{Computational Photography (ICCP), 2012 IEEE International Conference
  on}.\hskip 1em plus 0.5em minus 0.4em\relax IEEE, 2012, pp. 1--10.

\bibitem{CSLDS}
A.~C. Sankaranarayanan, P.~Turaga, R.~Baraniuk, and R.~Chellappa, ``Compressive
  acquisition of dynamic scenes,'' in \emph{Euro. Conf. Comp. Vision}, 2010.

\bibitem{donoho2006compressed}
D.~L. Donoho, ``Compressed sensing,'' \emph{{IEEE} Trans. Inf. Theory},
  vol.~52, no.~4, pp. 1289--1306, 2006.

\bibitem{candes2006near}
E.~J. Candes and T.~Tao, ``Near-optimal signal recovery from random
  projections: Universal encoding strategies?'' \emph{{IEEE} Trans. Inf.
  Theory}, vol.~52, no.~12, pp. 5406--5425, 2006.

\bibitem{tropp2007signal}
J.~A. Tropp and A.~C. Gilbert, ``Signal recovery from random measurements via
  orthogonal matching pursuit,'' \emph{{IEEE} Trans. Inf. Theory}, vol.~53,
  no.~12, pp. 4655--4666, 2007.

\bibitem{needell2009cosamp}
D.~Needell and J.~A. Tropp, ``Cosamp: Iterative signal recovery from incomplete
  and inaccurate samples,'' \emph{Applied and Computational Harmonic Analysis},
  vol.~26, no.~3, pp. 301--321, 2009.

\bibitem{EPFL}
V.~Thirumalai and P.~Frossard, ``Correlation estimation from compressed
  images,'' \emph{J. Visual Communication and Image Representation}, vol.~24,
  no.~6, pp. 649--660, 2013.

\bibitem{Rectex}
{K. Kulkarni and P. Turaga}, ``Recurrence textures for activity recognition
  using compressive cameras,'' in \emph{{IEEE} Conf. Image Process.}, 2012.

\bibitem{Davenport1}
{M. A. Davenport, M. F. Duarte, M. B. Wakin, J. N. Laska, D. Takhar, K. F.
  Kelly and R. G. Baraniuk}, ``The smashed filter for compressive
  classification and target recognition,'' pp. 6498--6499, 2007.

\bibitem{CLearning}
{R. Calderbank, S. Jafarpour and R. Schapire}, ``Compressed learning: Universal
  sparse dimensionality reduction and learning in the measurement domain,'' in
  \emph{Preprint}, 2009.

\bibitem{kulkarni2015reconstruction}
K.~Kulkarni and P.~Turaga, ``Reconstruction-free action inference from
  compressive imagers,'' \emph{{IEEE} Trans. Pattern Anal. Mach. Intell.},
  vol.~PP, no.~99, 2015.

\bibitem{lohit2015reconstruction}
S.~Lohit, K.~Kulkarni, P.~Turaga, J.~Wang, and A.~Sankaranarayanan,
  ``Reconstruction-free inference on compressive measurements,'' in
  \emph{Proceedings of the IEEE Conference on Computer Vision and Pattern
  Recognition Workshops}, 2015, pp. 16--24.

\bibitem{viola2004robust}
P.~Viola and M.~J. Jones, ``Robust real-time face detection,'' \emph{Intl. J.
  Comp. Vision}, vol.~57, no.~2, pp. 137--154, 2004.

\bibitem{dollar2010fastest}
P.~Doll{\'a}r, S.~Belongie, and P.~Perona, ``The fastest pedestrian detector in
  the west.'' in \emph{British Machine Vision Conf.}, vol.~2, no.~3, 2010,
  p.~7.

\bibitem{grabner2006real}
H.~Grabner, M.~Grabner, and H.~Bischof, ``Real-time tracking via on-line
  boosting.'' in \emph{British Machine Vision Conf.}, 2006.

\bibitem{zhang2012real}
K.~Zhang, L.~Zhang, and M.-H. Yang, ``Real-time compressive tracking,'' in
  \emph{Euro. Conf. Comp. Vision}, 2012, pp. 864--877.

\bibitem{babenko2011robust}
B.~Babenko, M.-H. Yang, and S.~Belongie, ``Robust object tracking with online
  multiple instance learning,'' \emph{{IEEE} Trans. Pattern Anal. Mach.
  Intell.}, vol.~33, no.~8, pp. 1619--1632, 2011.

\bibitem{kalal2010pn}
Z.~Kalal, J.~Matas, and K.~Mikolajczyk, ``Pn learning: Bootstrapping binary
  classifiers by structural constraints,'' in \emph{{IEEE} Conf. Comp. Vision
  and Pattern Recog}, 2010, pp. 49--56.

\bibitem{ramakanth2013seamseg}
S.~Ramakanth and R.~Babu, ``Seamseg: Video object segmentation using patch
  seams,'' in \emph{{IEEE} Conf. Comp. Vision and Pattern Recog}, 2013, pp.
  376--383.

\bibitem{duarte2009learning}
J.~M. Duarte-Carvajalino and G.~Sapiro, ``Learning to sense sparse signals:
  Simultaneous sensing matrix and sparsifying dictionary optimization,''
  \emph{IEEE Transactions on Image Processing}, vol.~18, no.~7, pp. 1395--1408,
  2009.

\bibitem{goldstein2013stone}
T.~Goldstein, L.~Xu, K.~F. Kelly, and R.~Baraniuk, ``The stone transform:
  Multi-resolution image enhancement and real-time compressive video,''
  \emph{arXiv preprint arXiv:1311.3405}, 2013.

\bibitem{chang2009informative}
H.~S. Chang, Y.~Weiss, and W.~T. Freeman, ``Informative sensing,'' \emph{arXiv
  preprint arXiv:0901.4275}, 2009.

\bibitem{Davenport}
{M. A. Davenport, M. F. Duarte, M. B. Wakin, J. N. Laska, D. Takhar, K. F.
  Kelly and R. G. Baraniuk}, ``The smashed filter for compressive
  classification and target recognition,'' \emph{Computat. Imag. V}, vol. 6498,
  pp. 142--153, 2007.

\bibitem{andrews1974scale}
D.~F. Andrews and C.~L. Mallows, ``Scale mixtures of normal distributions,''
  \emph{Journal of the Royal Statistical Society. Series B (Methodological)},
  pp. 99--102, 1974.

\bibitem{wainwright1999scale}
M.~J. Wainwright and E.~P. Simoncelli, ``Scale mixtures of gaussians and the
  statistics of natural images.'' in \emph{Adv. Neural Inf. Proc. Sys.}, 1999,
  pp. 855--861.

\bibitem{lyu2009modeling}
S.~Lyu and E.~P. Simoncelli, ``Modeling multiscale subbands of photographic
  images with fields of gaussian scale mixtures,'' \emph{{IEEE} Trans. Pattern
  Anal. Mach. Intell.}, vol.~31, no.~4, pp. 693--706, 2009.

\bibitem{mallat1989theory}
S.~G. Mallat, ``A theory for multiresolution signal decomposition: the wavelet
  representation,'' \emph{{IEEE} Trans. Pattern Anal. Mach. Intell.}, vol.~11,
  no.~7, pp. 674--693, 1989.

\bibitem{frahm2004generalized}
G.~Frahm, ``Generalized elliptical distributions: theory and applications,''
  Ph.D. dissertation, Universit{\"a}t zu K{\"o}ln, 2004.

\bibitem{candes2013phaselift}
E.~J. Candes, T.~Strohmer, and V.~Voroninski, ``Phaselift: Exact and stable
  signal recovery from magnitude measurements via convex programming,''
  \emph{Communications on Pure and Applied Mathematics}, vol.~66, no.~8, pp.
  1241--1274, 2013.

\bibitem{hegdenumax}
C.~Hegde, A.~C. Sankaranarayanan, W.~Yin, and R.~G. Baraniuk, ``Numax: A convex
  approach for learning near-isometric linear embeddings.''

\bibitem{liu2012implementable}
Y.-J. Liu, D.~Sun, and K.-C. Toh, ``An implementable proximal point algorithmic
  framework for nuclear norm minimization,'' \emph{Mathematical programming},
  vol. 133, no. 1-2, pp. 399--436, 2012.

\bibitem{cai2010singular}
J.-F. Cai, E.~J. Cand{\`e}s, and Z.~Shen, ``A singular value thresholding
  algorithm for matrix completion,'' \emph{SIAM Journal on Optimization},
  vol.~20, no.~4, pp. 1956--1982, 2010.

\bibitem{fukushima2002smoothing}
M.~Fukushima, Z.-Q. Luo, and P.~Tseng, ``Smoothing functions for
  second-order-cone complementarity problems,'' \emph{SIAM Journal on
  optimization}, vol.~12, no.~2, pp. 436--460, 2002.

\bibitem{pascal-voc-2007}
M.~Everingham, L.~Van~Gool, C.~K.~I. Williams, J.~Winn, and A.~Zisserman, ``The
  {PASCAL} {V}isual {O}bject {C}lasses {C}hallenge 2007 {(VOC2007)}
  {R}esults,''
  http://www.pascal-network.org/challenges/VOC/voc2007/workshop/index.html.

\bibitem{wu2013online}
Y.~Wu, J.~Lim, and M.-H. Yang, ``Online object tracking: A benchmark,'' in
  \emph{{IEEE} Conf. Comp. Vision and Pattern Recog}.\hskip 1em plus 0.5em
  minus 0.4em\relax IEEE, 2013, pp. 2411--2418.

\bibitem{hare2011struck}
S.~Hare, A.~Saffari, and P.~H. Torr, ``Struck: Structured output tracking with
  kernels,'' in \emph{{IEEE} Intl. Conf. Comp. Vision.}\hskip 1em plus 0.5em
  minus 0.4em\relax IEEE, 2011, pp. 263--270.

\bibitem{henriques2012exploiting}
J.~F. Henriques, R.~Caseiro, P.~Martins, and J.~Batista, ``Exploiting the
  circulant structure of tracking-by-detection with kernels,'' in \emph{Euro.
  Conf. Comp. Vision}, 2012, pp. 702--715.

\bibitem{zhang2014fast}
K.~Zhang, L.~Zhang, and M.~Yang, ``Fast compressive tracking,'' \emph{{IEEE}
  Trans. Pattern Anal. Mach. Intell.}, vol.~36, no.~10, pp. 2002--2015, 2014.

\bibitem{dong2014learning}
C.~Dong, C.~C. Loy, K.~He, and X.~Tang, ``Learning a deep convolutional network
  for image super-resolution,'' in \emph{Euro. Conf. Comp. Vision}.\hskip 1em
  plus 0.5em minus 0.4em\relax Springer, 2014, pp. 184--199.

\end{thebibliography}

		
		
		

		%

		\begin{IEEEbiography}{Kuldeep Kulkarni}
			
		\end{IEEEbiography}
		
		\begin{IEEEbiography}{Pavan Turaga}
			
		\end{IEEEbiography}
		
		
		

	\end{document}